\pdfoutput=1

\documentclass[11pt]{article}

\usepackage[final]{acl}

\usepackage{times}
\usepackage{latexsym}

\usepackage[T1]{fontenc}

\usepackage[utf8]{inputenc}

\usepackage{microtype}
\usepackage{booktabs}
\usepackage{kotex}
\usepackage{bbm}
\usepackage{algorithm}
\usepackage{url}
\usepackage{soul}
\usepackage{verbatim}
\usepackage{enumitem}
\usepackage{algorithm}
\usepackage{algorithmic}
\usepackage{amsmath, amssymb}       
\usepackage{mathtools}              
\usepackage{bm} 
\usepackage{subcaption}
\usepackage{multirow}
\usepackage{lipsum}
\usepackage{booktabs}
\usepackage{multirow}
\usepackage{booktabs}
\usepackage{inconsolata}
\usepackage[table]{xcolor}
\usepackage{graphicx}
\usepackage{caption}
\usepackage{tabularx}   
\usepackage{makecell}   
\usepackage{booktabs}   
\usepackage{booktabs,tabularx,siunitx}
\usepackage[x11names]{xcolor} 
\usepackage{booktabs}
\usepackage{multirow}
\usepackage{graphicx}
\usepackage{colortbl}
\usepackage{xcolor}
\usepackage{pifont}
\usepackage{amsmath, amssymb, amsfonts}

\newcommand{\km}[1]{\textcolor{black}{#1}}
\newcommand{\hj}[1]{\textcolor{black}{#1}}
\newcommand{\hh}[1]{\textcolor{black}{#1}}
\newcommand{\yr}[1]{\textcolor{black}{#1}}
\newcommand{\yj}[1]{\textcolor{black}{#1}}

\definecolor{myred}{HTML}{D9705C}
\definecolor{myblue}{HTML}{8FBFF0}
\definecolor{mygray}{HTML}{C2C5C8}

\newcommand{\nj}[1]{\textcolor{black}{#1}}

\definecolor{latinred}{HTML}{F4817B}
\definecolor{nonlatinblue}{HTML}{A5B3DC}

\title{\textbf{$\boldsymbol{\mu^2}$}-Bench: A Multilingual Machine Unlearning Benchmark}

\author{
  Kyomin Hwang$^{1*}$ \quad
  Hyeonjin Kim$^{1*}$ \quad
  Hyunho Lee$^{2*}$ \quad
  Yearim Kim$^{1*}$ \quad
  Yeji Song$^{1}$ \quad
  Nojun Kwak$^{1,2\dagger}$ \\[0.4em]
  $^1$GSCST, Seoul National University \\
  $^2$AIIS, Seoul National University \\
  \texttt{\{kyomin98, peaceful1, hhlee822, yerim1656, ldynx, nojunk\}@snu.ac.kr}
}

\begin{document}
\maketitle
\begin{abstract}

\km{Undesired information such as harmful content and private data propagates through Multilingual Large Language Models (LLMs) via direct training and indirect cross-linguistic spread. Multilingual Machine Unlearning (MMU) aims to remove such information, yet its evaluation remains underexplored, leaving unclear whether unlearning truly eliminates \nj{target} knowledge across all languages. To bridge this gap, we introduce $\boldsymbol{\mu^2}$-Bench, an MMU benchmark that simulates the full pipeline of memorization, unlearning, and evaluation across diverse languages. It 1) spans a broad set of languages, 2) evaluates on both training and hold-out languages, and 3) assesses knowledge as dispersed across multiple languages. We show that successful MMU requires methods that reflect multilingual characteristics, and conduct analysis to provide deeper insights into MMU.}

\end{abstract}

\begin{figure}[t]
    \centering
    \includegraphics[width=0.85\linewidth]{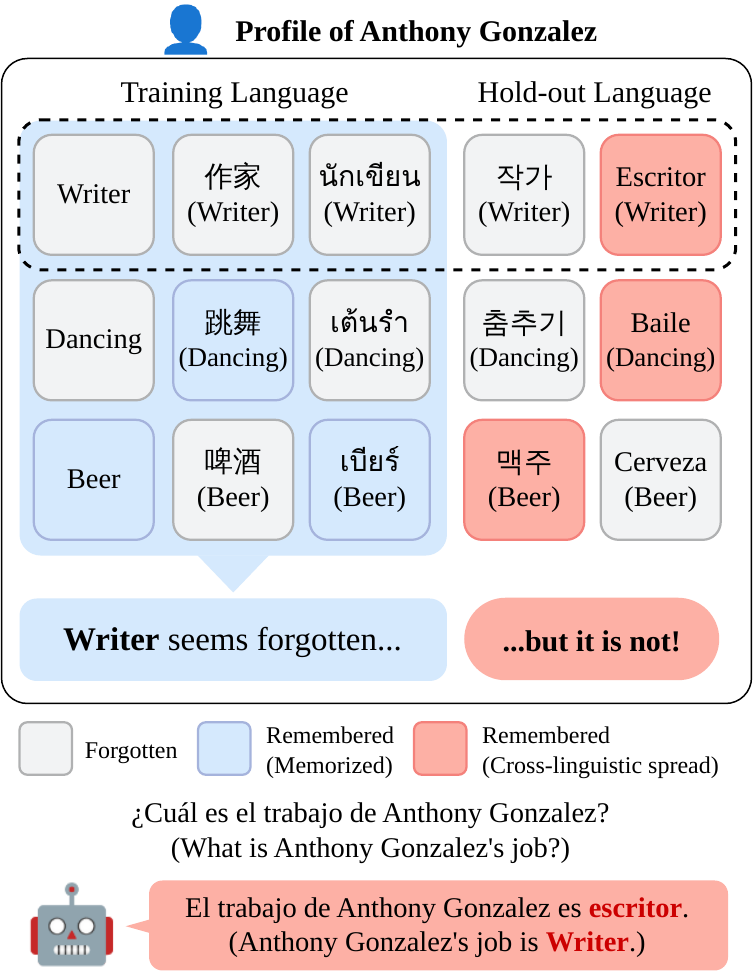}
    \caption{\km{Illustration of knowledge after unlearning. Columns correspond to languages, and rows correspond to knowledge instances expressed across multiple languages. \textcolor{mygray}{Gray} cells indicate knowledge forgotten in the corresponding language, whereas colored cells indicate the retained ones, either acquired directly via training (\textcolor{myblue}{blue}) or indirectly via cross-linguistic spread (\textcolor{myred}{red}).}}
    \vspace{-5mm}
    \label{fig:setting}
\end{figure}

\begin{figure*}[t]
    \centering
    \includegraphics[width=0.9\linewidth]{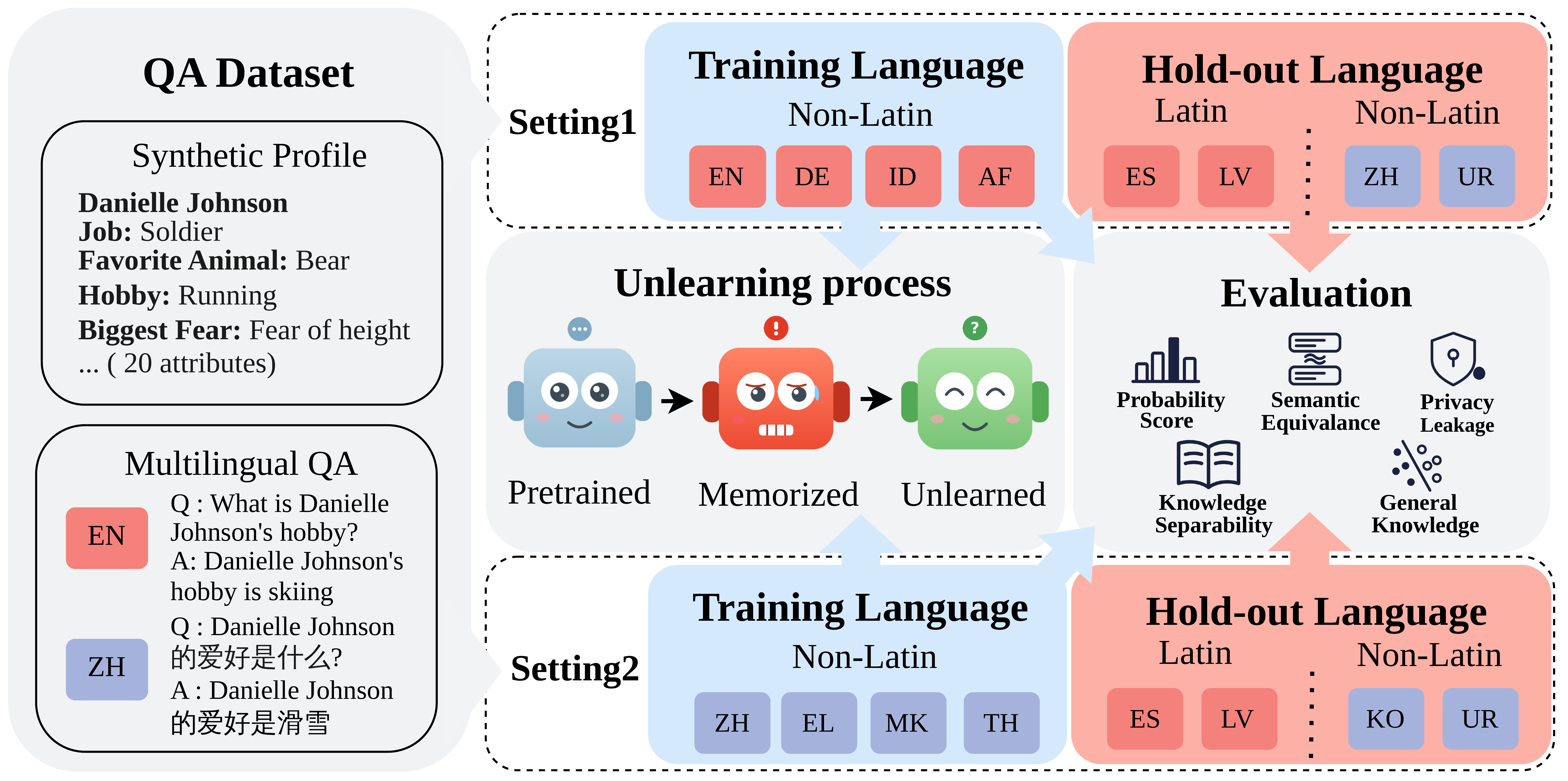}
    \vspace{-2mm}
\caption{\hh{Overview of the \nj{$\boldsymbol{\mu^2}$}-Bench: (Left) a multilingual dataset is constructed, (Middle) memorization and unlearning are performed using only the training languages of each setting, (Right) and the unlearned model is evaluated on both the training and hold-out languages. Latin and non-Latin languages are denoted in \textcolor{myred}{red} and \textcolor{myblue}{blue}.}}
\vspace{-3mm}
\label{fig:pipeline}
\end{figure*}

\section{Introduction}

How do LLMs acquire knowledge? Modern LLMs are typically pretrained on large multilingual corpora spanning various languages. As a result, LLMs gain the capability to embed knowledge multilingually. When further fine-tuned, each knowledge instance within the fine-tuning set becomes embedded across \nj{both the} training languages in which the knowledge was directly expressed in the fine-tuning set and \nj{the} hold-out languages in which it was not. This spread of knowledge into hold-out languages is a phenomenon called cross-linguistic spread~\cite{cla,linguafranca}. The same applies to \nj{undesired} knowledge: \nj{harmful or private content} acquired by a model becomes embedded across both training and hold-out languages~\cite{lau}. Under this scenario, the need to remove such knowledge has motivated the research on Multilingual Machine Unlearning (MMU)~\cite{lingtea}.

\yj{However, the evaluation of MMU remains underexplored.} \km{MMU should be evaluated from three perspectives: 1) whether the evaluation accounts for linguistic diversity, such as differences in resource level, 2) whether forgetting occurs not only in the training languages but also in the hold-out languages, and 3) whether the knowledge \nj{spanning multiple languages} itself is removed as a whole, rather than being assessed separately in a language-wise manner. \km{For instance}, as in the dotted \nj{first row} of Figure~\ref{fig:setting}, a model may forget target knowledge (\textit{e.g.,} writer) in training languages but continue to retrieve it in hold-out languages~\cite{knowledge}.
In such cases, evaluating unlearning solely in the training language\yr{, and} in a language-wise manner is insufficient.
However, existing benchmarks address these perspectives only partially, thereby failing to capture such cases and leaving the effectiveness of MMU inadequately measured~\cite{fame, amnesia, multitofu}.}

In this paper, we present \nj{\textbf{$\boldsymbol{\mu^2}$-Bench (\textbf{MU}ltilingual \textbf{M}achine \textbf{U}nlearning Benchmark)}}, a benchmark for MMU that addresses these limitations in three dimensions: dataset, framework, and evaluation. 
First, we construct a multilingual dataset spanning 12 languages, considering both scripts and resource-levels, which enables a comprehensive analysis across diverse linguistic properties. Second, \hj{as illustrated in Figure~\ref{fig:pipeline},} we propose a simulation framework that explicitly accounts for cross-linguistic spread, allowing us to evaluate whether \yr{target} knowledge remains accessible in hold-out languages \yr{after unlearning}. \yj{Third, we apply knowledge-wise evaluation metrics to our benchmark, enabling us to assess MMU \km{in a language-agnostic manner}.} Together, these components provide a more rigorous evaluation of MMU.

To this end, we present a comprehensive set of experiments conducted with \nj{$\boldsymbol{\mu^2}$}-Bench. Our experiments show that MMU requires methods that explicitly account for the multilingual nature of the task. Furthermore, we provide a detailed analysis of unlearning behavior across training and hold-out languages, examined along the dimensions of script and resource level. Moreover, we show that MMU can induce code-mixed outputs as an unintended side effect, and discuss how this phenomenon complicates the evaluation of unlearning. \hh{Collectively, these contributions establish both a benchmark for evaluating MMU and a set of analyses that deepen the understanding of MMU.} 
\section{Related Work}



\subsection{\yr{Cross-linguistic Spread} in LLMs}

\yr{Modern LLMs are trained in multilingual environments, which in turn allow \km{undesired} information to be embedded across diverse languages}~\cite{lau, knowledge}. \hj{LLMs possess the ability to retrieve knowledge from languages not explicitly \nj{used} in training, a phenomenon attributed to cross-linguistic spread~\cite{pires2019multilingual, conneau2020emerging, shaham2024multilingual}. \hh{This ability allows undesired information to propagate to hold-out languages, thereby posing a new safety challenge.}} 
\subsection{Multilingual Machine Unlearning (MMU)} 
\hj{MMU} aims to remove \nj{undesired} information, expressed in multiple languages, from a trained model~\cite{lingtea,multi_erasure}. \hh{\citet{lingtea} show that unlearning a multilingual LLM on an English-only dataset is insufficient, accordingly proposing a method to tackle this.} \hj{\citet{lau} extend their intuition and show that the knowledge acquired through cross-linguistic spread persists after English-only unlearning.} \hj{Despite such growing \nj{body of literature}, evaluation framework for MMU remains \nj{under}explored.}

\subsection{\hj{Benchmarking and Evaluating MMU}}

There have been few efforts to benchmark and evaluate MMU. For example, \citet{multitofu} as well as \citet{amnesia} translate the English unlearning benchmark TOFU~\cite{tofu} into multiple languages. \citet{fame} follow TOFU to construct a synthetic MMU benchmark centered on fictional actors. However, these benchmarks have two limitations in evaluation: 1) they are confined to language-wise evaluation, and 2) they do not evaluate the knowledge acquired through cross-linguistic spread. As a result, they offer limited evidence as to whether a knowledge instance has been removed across all of its multilingual expressions. \citet{knowledge} partially address this limitation by proposing an evaluation framework that includes two hold-out languages, together with a metric for measuring unlearning across languages. However, the hold-out languages in their work are limited to Latin-script languages, insufficient for a comprehensive analysis. In this paper, we address the two limitations more extensively, and provide a benchmark that offers thorough simulation of MMU, encompassing memorization, unlearning, and evaluation.

\section{Problem Formulation}

\label{sec:formulation}

\km{In this section, we introduce the notion of \yr{knowledge}, training languages and hold-out languages. \hj{Subsequent sections of the paper build on these formulations to examine the dynamics in MMU.}} 

\subsection{\yr{Multilingual Knowledge Instances}}

\yr{Following \citet{knowledge}, we view a knowledge instance as a language-agnostic unit that can be expressed in multiple languages.} \yr{Let $\mathbb{L}$ denote the set of languages considered in the benchmark.} \hj{The $i$-th knowledge \yr{instance} $k_{i,\mathbb{L}}$ is defined as:} 

\vspace{-1mm}
\begin{equation}
    k_{i,\mathbb{L}}=(k_{i,\ell})_{\ell \in \mathbb{L}},
    \vspace{-1mm}
\end{equation}
\noindent\km{where 
$k_{i,\ell}$ is the knowledge expressed in each language $\ell$.}
\yr{In our benchmark, each $k_{i,\ell}$ corresponds to language-specific knowledge that expresses the same underlying fact as $k_{i,\mathbb{L}}$.}
\yr{The complete multilingual knowledge set can therefore be viewed as a matrix whose rows correspond to knowledge instances and whose columns correspond to languages.} \yr{We partition the set of knowledge indices $\mathcal{I}$ into target (to be unlearned) and non-target (to be retained) subsets, denoted by $\mathcal{I}_\mathrm{T}$ and $\mathcal{I}_\mathrm{N}$ each.}

\vspace{-1mm}
$$
\begin{aligned}
    \mathcal{D}_{f} &= \{\, k_{i,\mathbb{L}} \mid i \in \mathcal{I}_\mathrm{T}\,\}, \\
    \mathcal{D}_r &= \{\, k_{j,\mathbb{L}}  \mid j \in \mathcal{I}_\mathrm{N}\,\},
\end{aligned}
$$
\vspace{-1mm}

\noindent \km{where $\mathcal{D}_f$ and $\mathcal{D}_r$ denote the forget and the retain sets, respectively.} The full dataset $\mathcal{D} = \mathcal{D}_f \cup \mathcal{D}_r$ can be viewed as an $|\mathcal{I}| \times |\mathbb{L}|$ matrix. \km{Building on this view, we propose an MMU benchmark that considers the knowledge-wise perspective.} \km{In Figure~\ref{fig:setting}, \nj{each row corresponds to a knowledge instance}.}

\subsection{Language Set Partitioning}
\label{sec:knowledge_acquisition}

Language-specific knowledge $k_{i, \ell}$ can be acquired through two distinct pathways: \yr{1) direct acquisition, where the model \hj{has been exposed} to $k_{i, \ell}$ during memorization, and 2) indirect acquisition, where the model has not \hj{been directly exposed} but can still access the underlying knowledge through cross-linguistic spread.}
\yr{
To formalize this distinction, we partition the language set $\mathbb{L}$ into training languages $\mathbb{L}^\mathrm{train}$ and hold-out languages $\mathbb{L}^\mathrm{hold}$:}
\begin{equation}
    \mathbb{L}
    =
    \mathbb{L}^{\mathrm{train}}
    \cup
    \mathbb{L}^{\mathrm{hold}},
    \quad
    \mathbb{L}^{\mathrm{train}}
    \cap
    \mathbb{L}^{\mathrm{hold}}
    =
    \emptyset .
\end{equation}

Languages in $\mathbb{L}^{\mathrm{train}}$ are used during memorization and unlearning, whereas languages in  $\mathbb{L}^{\mathrm{hold}}$ are excluded from both stages and reserved for evaluation. \km{In our \nj{$\boldsymbol{\mu^2}$}-Bench, we evaluate MMU across both splits of languages.}

\section{\nj{\textit{µ}$^2$}-Bench: A Multilingual Machine Unlearning Benchmark}

\km{In this section, we introduce \textbf{\nj{$\boldsymbol{\mu^2}$}-Bench}, a \textbf{MU}ltilingual \textbf{M}achine \textbf{U}nlearning Benchmark \yr{for evaluating} 
MMU.} \yr{\hj{$\boldsymbol{\mu^2}$-Bench consists of three components: 1) a synthetic multilingual QA dataset, 2) controlled MMU framework including dataset splits and unlearning configurations, and 3) an evaluation protocol that includes both training and hold-out languages.} The benchmark is designed to assess whether target knowledge is removed across languages, even where it was not directly observed during memorization \km{and} unlearning.}

\subsection{Dataset Construction}
\label{sec:dataset_construction}

\begin{figure}[t]
    \centering
    \includegraphics[width=0.88\linewidth]{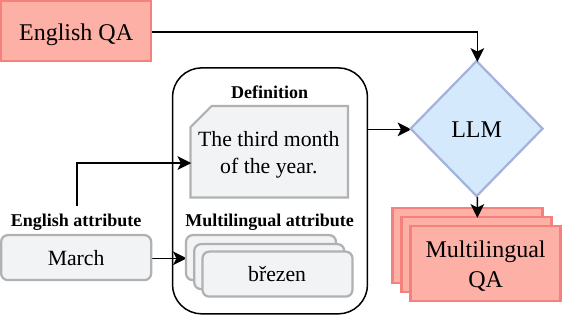}
    \caption{\hj{The construction of the translation prompt. To guarantee the translation quality, the prompt includes the definition of each English attribute along with its translation into the target language.}}
    \vspace{-3mm}
    \label{fig:pipeline_trans}
\end{figure}

\hj{We construct a multilingual parallel dataset of 60,000 question-answer (QA) pairs across 12 languages, comprising 5,000 knowledge instances across 250 fictitious profiles.}
Each synthetic profile is characterized by 20 attributes that yield a corresponding QA pair.
The overall construction pipeline is in three stages: 1) generating synthetic profiles in English, 2) constructing English QA datasets from each profile, 3) translating the QA datasets into multiple target languages, and validating the translations through back-translation and human-in-the-loop refinement. \km{Detailed additional explanations are provided in the Appendix~\ref{sec:dataset_detail}.}

\paragraph{Synthetic English Profiles} We constructed a total of 250 distinct synthetic profiles, \nj{consisting} of a randomly generated name and 20 independent attributes. The names were generated in English using the Faker library~\cite{faker}, and the attributes were sampled from predefined value pools to characterize each profile. All entries were verified by human annotators to ensure \nj{no} ambiguity. 


\paragraph{English QA Dataset} 
Based on the constructed profiles, we generated one English QA pair per attribute, yielding a total of 5,000 pairs (250 $\times$ 20). Specifically, we adopted a template-based approach in which attribute values were mapped to predefined templates to produce the initial QA pairs. The initial set was subsequently refined using the GPT-5.4-mini API~\cite{gpt5}. Human annotators then reviewed the pairs to ensure that each question precisely targeted its intended attribute. 

\paragraph{Multilingual QA Dataset} 
To extend the English QA dataset into a multilingual setting, we translated the 5,000 English QA pairs into 11 target languages using the GPT-5.4-mini API~\cite{gpt5}, resulting in a multilingual QA dataset of 60,000 instances (12 $\times$ 5,000) including English. \km{The twelve languages are English (\textsc{en}), German (\textsc{de}), Indonesian (\textsc{id}), Afrikaans (\textsc{af}), Spanish (\textsc{es}), Latvian (\textsc{lv}), Chinese (\textsc{zh}), Korean (\textsc{ko}), Urdu (\textsc{ur}), Greek (\textsc{el}), Macedonian (\textsc{mk}), Thai (\textsc{th}).} The translation proceeded in four steps. First, the English values in the predefined value pools were translated into each target language. Second, the English values were annotated with their corresponding definition to preserve its intended meaning. Third, the English QA pairs, together with the translated attributes and their definitions, were concatenated into a single prompt for machine translation. Finally, the translated QA sets were refined through a human-in-the-loop process. For the final refinement process, we employed a back-translation-based quality control pipeline: each translated instance was back-translated into English \km{with the Google Translation API}~\cite{google-translate}, and the semantic consistency between the original and the back-translated sentence was verified by human \km{annotators}. The overall construction pipeline of the multilingual QA dataset is illustrated in Figure~\ref{fig:pipeline_trans}. 

\subsection{\hj{MMU Simulation Framework}}
\label{sec:setting}
\begin{table}[t]
\centering
\small
\begin{tabular}{lccc}
\toprule
\textbf{Setting} & \textbf{Script} & \textbf{Training} & \textbf{Hold-out} \\
\midrule
\multirow{2}{*}{Setting 1}
 & Latin     & \textbf{EN, DE}, ID, AF & \textbf{ES}, LV \\
 & Non-Latin & ---                     & \textbf{ZH}, UR \\
\midrule
\multirow{2}{*}{Setting 2}
 & Latin     & ---                     & \textbf{EN}, LV \\
 & Non-Latin & \textbf{ZH, EL}, MK, TH & \textbf{KO}, UR \\
\bottomrule
\end{tabular}
\caption{\yr{Language} configurations \yr{used in \textbf{\nj{$\boldsymbol{\mu^2}$}-Bench}} across two evaluation settings, organized by writing script (Latin vs. non-Latin) and resource level (High vs. Low). \textbf{Bold} indicates high-resource languages.} 
\label{tab:setting}
\end{table}
We present an overall pipeline of \nj{$\boldsymbol{\mu^2}$}-Bench \yr{in three stages}: memorization, unlearning, and evaluation.

\paragraph{Memorization}
\yr{The model is fine-tuned on the memorization set expressed only in the training languages, \km{$k_{i,\mathbb{L}^\mathrm{train}}$}, where $i \in \mathcal{I}_\mathrm{T}\cup \mathcal{I}_\mathrm{N}$ denotes the knowledge index.} \hj{For controlled observation, we design two settings with different training languages as \yr{summarized in Table~\ref{tab:setting}. In Setting 1, the training languages are all Latin script languages, whereas in Setting 2, they are all non-Latin script languages. In both settings, the hold-out language set contains two Latin script and two non-Latin script languages, and both the training and hold-out sets are each balanced with two high-resource and two low-resource languages.}} \km{Within this setup, 200 of the 250 profiles \hj{constructed} are used for memorization, contributing 16,000 QA pairs in total across the four training languages. The remaining 50 profiles are reserved as non-members for the privacy leakage evaluation.}


\paragraph{Unlearning}

\km{Unlearning is applied to the target knowledge expressed in the training languages, $k_{i,\mathbb{L}^\mathrm{train}}$. We conduct unlearning under three settings \yr{with} the forget ratio: $1\%$ ($p1$), $3\%$ ($p3$), and $5\%$ ($p5$). For each setting, the corresponding percentage of the 200 individuals used during memorization is designated as the forget set $\mathcal{D}_f$, while the remaining constitute the retain set $\mathcal{D}_r$.}


\paragraph{Evaluation}

\km{Using the unlearned model, we assess how effectively the target knowledge has been removed, reporting results separately for the training and the hold-out languages. This separation allows us to analyze how each unlearning method behaves within each language group. The hold-out languages used for evaluation in each setting are listed in Table~\ref{tab:setting}.} \km{To this end, we provide an evaluation pipeline that considers not only the training languages but also the hold-out languages overlooked by prior benchmarks. This enables controlled evaluation across diverse settings by varying the memorization languages.}

\subsection{Evaluation Protocol} 
\label{subsec:evaluation-protocol}
\hj{
The evaluation is conducted using the constructed QA pairs, denoted as $k_{i,\ell}=(q_{i,\ell}, a_{i,\ell})$ \km{with a slight abuse of notation}. As the proxy for ideal unlearning, we employ a model trained exclusively on non-target knowledge, denoted as $f_R$. The reference model $f_R$ is trained separately for Setting 1 and Setting 2, using both training and hold-out languages within each setting. \km{Since $f_R$ is trained exclusively on the non-target knowledge while excluding the target knowledge, a successfully unlearned model should closely resemble $f_R$. Consequently, $f_R$ can serve as the gold standard.} \nj{$\boldsymbol{\mu^2}$}-Bench measures MMU from three perspectives: aggregated forget quality (FQ1-3), separability quality \yr{(SQ)} between the target and the non-target knowledge\, and the maintenance of general utility \yr{quality}~(UQ).}

\paragraph{FQ1: Probability Score}
\hj{The Probability score is calculated by averaging the probability of the ground-truth answer when given the questions:}

\vspace{-2mm}
\begin{equation}
    \mathrm{PS}=\frac{1}{|\mathcal{I}|} \sum_{i \in \mathcal{I}} \frac{1}{|\mathbb{L}|}\sum_{\ell \in \mathbb{L}}\mathcal{P}(a_{i, \ell} | q_{i, \ell})^{{1}/{|a_{i, \ell}|_\mathrm{tok}}}. 
\end{equation}

\noindent \hj{Here, $|a_{i, \ell}|_\mathrm{tok}$ represents the number of tokens in $a_{i, \ell}$. A higher score denotes stronger memorization and less forgetting of the ground truth.}

\paragraph{FQ2: Semantic Equivalence Score}
\hj{The Semantic Equivalence score validates whether the output generated by a model $f$ matches the ground-truth:}

\vspace{-2mm}
\begin{equation}
    \mathrm{SE}=\frac{1}{|\mathcal{I}|}\sum_{i \in \mathcal{I}}\frac{1}{|\mathbb{L}|}\sum_{\ell \in \mathbb{L}}\mathbb{I}(\mathrm{LLM}(f(q_{i,\ell}),a_{i,\ell})).
\end{equation}

\noindent \hj{In this paper, we employ the GPT-4o-mini API with greedy decoding to verify equivalence.} \km{The prompt used for evaluation can be found in Figure~\ref{fig:judge_prompt}.}

\paragraph{FQ3: Privacy Leakage}

\hj{The Membership inference attack (MIA) \yr{is} adopted by MUSE~\cite{muse} in the form of Privacy Leakage:}

\begin{equation}
\small
    \text{PrivLeak}=\frac{\text{AUC}(f;\mathcal{D}_f, \mathcal{D}_h)-\text{AUC}(f_R;\mathcal{D}_f, \mathcal{D}_h)}{\text{AUC}(f_R;\mathcal{D}_f, \mathcal{D}_h)},
\end{equation}

\noindent \hj{where $\mathcal{D}_h$ denotes the 50 non-member profiles that were excluded from training. 
Privacy Leakage quantifies the extent to which a model $f$ resembles $f_R$ in distinguishing between non-member data and target knowledge. 
For this metric, the calculation \yr{is} performed by aggregating at the knowledge level.
An absolute value closer to $0$ indicates better resemblance, thus more successful unlearning. 
We follow the implementation of OpenUnlearning~\cite{openunlearning} and flip the sign of the value when reporting the result.} 


\paragraph{SQ: Knowledge Separability Score (KSS)}
\hj{We adopt KSS from \citet{knowledge}, and normalize it to obtain a more intuitive metric. KSS quantifies how well a model differentiates the knowledge instances $k_{i,\mathbb{L}}$ targeted for unlearning ($i \in \mathcal{I}_\mathrm{T}$) from those that are not ($i \in \mathcal{I}_\mathrm{N}$). For the calculation, the preceding work first defines the forgetting score of $i$-th knowledge as:}


\begin{equation}
    S_i= 1-\frac{1}{|\mathbb{L}|}\sum_{\ell \in \mathbb{L}}s(k_{i, \ell}),
\end{equation}

\noindent \hj{where $s(k_{i,\ell})$ can be either token probability $\mathcal{P}(a_{i,\ell}|q_{i,\ell})^{{1}/{|a_{i, \ell}|_\mathrm{tok}}}$ or semantic equivalence $\mathbb{I}(\mathrm{LLM}(f(q_{i,\ell}),a_{i,\ell}))$. From the distributions of $S_i$ collected from the target knowledge and the non-target knowledge, respectively, the Area Under the Receiver Operating Characteristic Curve ($\text{KSS}^\text{ROC}$) and the Precision-Recall Curve ($\text{KSS}^\text{PR}$) are computed.}
\hj{We normalize KSS as follows:}

\begin{equation}
    \text{KSS-Norm}=\frac{\text{KSS($f$)}-\text{KSS($f_R$)}}{\text{KSS($f_R$)}}.
\end{equation}

\noindent \hj{KSS-Norm quantifies how well a model $f$ resembles the reference model $f_R$ in distinguishing between the target and non-target knowledge. Below, KSS-Norm is abbreviated as KSS for conciseness.} 

\paragraph{UQ: General Knowledge}
\hh{To verify that unlearning has not degraded the general ability of an LLM, we measure the model's utility with TyDi QA~\cite{tydiqa}, complementing the retain quality measured on the retain set $\mathcal{D}_r$. TyDi QA reports performance with F1-score and EM score.}
\section{Experiments}


\subsection{Experimental Setup}
\label{sec:experimental_setup}

\paragraph{Baselines} \km{We compare four unlearning methods. We adopt \hj{three MU baselines into the multilingual setting}: Gradient Ascent (\textsc{GA})~\cite{ga}, Gradient Difference (\textsc{GAGDR})~\cite{tofu}, and Negative Preference Optimization (\textsc{NPO})~\cite{npo}. We additionally include LingTea~\cite{lingtea}, a method specifically designed for MMU. All of the experiments are conducted on a single A6000 GPU.}

\paragraph{Target Models} \km{We conduct unlearning experiments on two models: Gemma3-12B~\cite{gemma3} and Qwen2.5-7B-Instruct~\cite{qwen3}.}

\paragraph{Hyperparameter Search} 
\hj{We search for unlearning hyperparameters so that the Probability score of the retain set meets the predefined threshold across all methods. The threshold is set to 90\% of the Probability score of the fully memorized model, evaluated on $\{k_{i,\ell} \, | \, i\in \mathcal{I}_\mathrm{N}, \ell \in \mathbb{L}^\mathrm{train}\}$. Details and full hyperparameter configurations are reported in Appendix~\ref{sec:hyperparameter}.}

\paragraph{Metrics} 
\hh{We measure the overall degree of forgetting through the Probability \yr{Score} (\textsc{PS}) and Semantic Equivalence (\textsc{SE}) scores on the forget set $\mathcal{D}_f$, where a lower score indicates better forgetting. To capture separability between the target and non-target knowledge, we further report four normalized KSS variants, where a larger value indicates that the unlearned model separates the target knowledge from the non-target knowledge more effectively, measured relative to the reference model $f_R$. We additionally report the Privacy Leakage score (\textsc{PrivLeak}), for which a value closer to $0$ indicates that the unlearned model more closely resembles $f_R$. Finally, we verify that general ability is preserved using TyDi QA~\cite{tydiqa}, reporting the average of its F1 and EM scores as General Knowledge (\textsc{Gen Know.}).}

\subsection{Results}
\label{sec:results}

\begin{table*}[t]
\centering
\tiny
\setlength{\tabcolsep}{2pt}
\renewcommand{\arraystretch}{0.90}
\setlength{\extrarowheight}{0pt}
\setlength{\aboverulesep}{0pt}
\setlength{\belowrulesep}{0pt}
\resizebox{\textwidth}{!}{
\begin{tabular}{cccl |ccccccc| ccccccc| c}
\toprule
& & & & \multicolumn{7}{c}{\textbf{Train}} & \multicolumn{7}{c}{\textbf{Holdout}} & \multirow{2}{*}{\makecell{\textbf{Gen.}\\\textbf{Know.}}} \\
\cmidrule(lr){5-11} \cmidrule(lr){12-18}
Model & Set & $p$ & Method
& PS & SE & KSS$_{\text{prob}}^{\text{ROC}}$ & KSS$_{\text{prob}}^{\text{PR}}$ & KSS$_{\text{gen}}^{\text{ROC}}$ & KSS$_{\text{gen}}^{\text{PR}}$ & PrivLeak
& PS & SE & KSS$_{\text{prob}}^{\text{ROC}}$ & KSS$_{\text{prob}}^{\text{PR}}$ & KSS$_{\text{gen}}^{\text{ROC}}$ & KSS$_{\text{gen}}^{\text{PR}}$ & PrivLeak
& \\
\midrule
\multirow{30}{*}{\rotatebox{90}{Gemma3-12B}}
& \multirow{15}{*}{\rotatebox{90}{S1}} & \multirow{5}{*}{p1} & Mem     & 99.99 & 100.00 & -53.18 & -99.06 & -50.00 & -99.00 & -107.90 & 55.82 & 76.88 & -58.37 & -99.17 & -50.17 & -98.99 & -86.85 & 30.19 \\
&  &  & GA      & 72.44 & \underline{91.88} & \underline{-14.24} & -77.47 & -41.73 & -93.94 & -90.57 & 37.45 & \underline{68.75} & -42.05 & -97.21 & \textbf{-44.63} & \underline{-98.68} & -60.75 & 32.85 \\
&  &  & GAGDR   & 75.34 & 94.38 & -20.39 & -76.45 & -42.89 & -94.68 & -97.23 & 45.03 & 70.00 & -49.86 & -98.61 & \underline{-44.79} & -98.75 & -75.34 & 32.83 \\
&  &  & NPO     & \underline{47.89} & \textbf{83.12} & \textbf{-4.65} & \underline{-43.28} & \textbf{-33.69} & \textbf{-86.37} & \underline{-68.61} & \underline{26.09} & \textbf{67.50} & \textbf{-26.60} & \underline{-92.70} & -45.54 & \textbf{-98.59} & \underline{-41.32} & 32.49 \\
&  &  & LingTea & \textbf{45.32} & 91.88 & -20.23 & \textbf{-41.52} & \underline{-38.96} & \underline{-89.00} & \textbf{-1.07} & \textbf{25.92} & 73.12 & \underline{-28.49} & \textbf{-47.51} & -46.95 & -98.88 & \textbf{+9.18} & 30.84 \\
\cmidrule(lr){3-19}
&  & \multirow{5}{*}{p3} & Mem     & 99.99 & 100.00 & -49.32 & -96.14 & -48.94 & -96.87 & -111.15 & 54.25 & 80.00 & -56.74 & -97.49 & -50.60 & -96.96 & -82.18 & 30.19 \\
&  &  & GA      & 93.41 & 100.00 & -39.18 & \underline{-79.13} & -48.95 & -96.87 & -109.28 & 53.54 & 78.96 & -56.62 & -97.48 & -49.01 & -96.88 & -82.20 & 30.05 \\
&  &  & GAGDR   & \underline{79.84} & 100.00 & \underline{-25.42} & -82.99 & -49.08 & -96.87 & \underline{-106.20} & \underline{50.85} & \underline{78.33} & \underline{-55.97} & \underline{-97.44} & -49.36 & -96.91 & -82.16 & 30.90 \\
&  &  & NPO     & 87.83 & \underline{99.79} & -38.71 & -88.04 & \underline{-48.56} & \underline{-96.73} & -109.30 & 53.20 & 78.33 & -56.57 & -97.48 & \underline{-47.90} & \underline{-96.82} & \underline{-81.88} & 30.08 \\
&  &  & LingTea & \textbf{49.84} & \textbf{92.08} & \textbf{-23.55} & \textbf{-44.29} & \textbf{-38.86} & \textbf{-86.15} & \textbf{-30.58} & \textbf{28.27} & \textbf{71.04} & \textbf{-33.33} & \textbf{-59.80} & \textbf{-45.44} & \textbf{-96.58} & \textbf{-8.91} & 35.08 \\
\cmidrule(lr){3-19}
&  & \multirow{5}{*}{p5} & Mem     & 99.99 & 100.00 & -54.44 & -94.99 & -48.98 & -94.80 & -105.66 & 50.98 & 79.38 & -51.84 & -95.34 & -51.00 & -95.06 & -90.03 & 30.19 \\
&  &  & GA      & 87.48 & 100.00 & -43.17 & -91.85 & -49.09 & -94.80 & -104.35 & 49.68 & 77.75 & -51.55 & -95.29 & -49.34 & -94.91 & -89.16 & 30.29 \\
&  &  & GAGDR   & \underline{74.32} & \underline{97.50} & \underline{-32.01} & \underline{-68.19} & \underline{-46.11} & \underline{-92.14} & \underline{-87.25} & \textbf{36.80} & \underline{75.12} & \underline{-41.31} & \underline{-82.87} & -49.01 & \underline{-94.81} & \underline{-60.91} & 31.99 \\
&  &  & NPO     & 89.00 & 100.00 & -44.26 & -92.59 & -49.02 & -94.80 & -104.09 & 50.31 & 77.62 & -51.66 & -95.32 & \underline{-48.93} & -94.85 & -89.76 & 30.12 \\
&  &  & LingTea & \textbf{72.16} & \textbf{94.12} & \textbf{-31.98} & \textbf{-55.95} & \textbf{-41.11} & \textbf{-86.12} & \textbf{-68.16} & \underline{37.30} & \textbf{71.25} & \textbf{-39.50} & \textbf{-80.95} & \textbf{-44.39} & \textbf{-94.14} & \textbf{-57.09} & 31.46 \\
\cmidrule(lr){2-19}
& \multirow{15}{*}{\rotatebox{90}{S2}} & \multirow{5}{*}{p1} & Mem     & 99.99 & 100.00 & -53.51 & -99.06 & -50.00 & -99.00 & -115.38 & 50.09 & 78.75 & -58.47 & -99.19 & -53.42 & -99.08 & -106.75 & 23.12 \\
&  &  & GA      & 75.68 & 92.50 & -19.88 & -77.47 & -41.62 & -93.36 & -106.49 & 42.91 & 73.12 & -50.67 & -98.94 & -51.92 & -99.00 & -92.67 & 22.84 \\
&  &  & GAGDR   & \textbf{45.07} & \textbf{74.38} & \underline{-12.38} & -40.97 & -27.29 & \underline{-77.69} & \textbf{-30.11} & \underline{29.78} & \underline{55.62} & \underline{-33.85} & \underline{-96.42} & \underline{-36.04} & \underline{-98.43} & \underline{-58.82} & 23.27 \\
&  &  & NPO     & 55.04 & 80.00 & -14.17 & \underline{-39.63} & \underline{-26.63} & -81.90 & \underline{-70.04} & 42.69 & 69.38 & -50.97 & -98.91 & -47.61 & -98.87 & -91.04 & 23.05 \\
&  &  & LingTea & \underline{54.23} & \underline{78.75} & \textbf{-5.06} & \textbf{-39.52} & \textbf{-23.66} & \textbf{-75.32} & -72.67 & \textbf{16.85} & \textbf{54.37} & \textbf{-12.74} & \textbf{-73.43} & \textbf{-32.70} & \textbf{-98.31} & \textbf{-13.53} & 22.42 \\
\cmidrule(lr){3-19}
&  & \multirow{5}{*}{p3} & Mem     & 99.99 & 100.00 & -49.57 & -96.86 & -48.94 & -96.87 & -100.46 & 49.91 & 80.00 & -58.53 & -97.62 & -54.52 & -97.26 & -84.92 & 23.12 \\
&  &  & GA      & 87.65 & \underline{98.33} & -45.19 & -92.91 & -47.94 & -96.79 & -98.50 & 49.70 & 79.58 & -58.80 & -97.62 & -54.67 & -97.24 & -84.54 & 23.08 \\
&  &  & GAGDR   & \underline{75.50} & 98.33 & \textbf{-26.91} & \textbf{-58.71} & \underline{-46.51} & \underline{-94.53} & \underline{-77.60} & \underline{42.37} & \underline{75.21} & \underline{-46.54} & \underline{-96.04} & \textbf{-50.96} & \underline{-97.06} & \underline{-68.34} & 22.02 \\
&  &  & NPO     & 87.80 & 98.33 & -44.38 & -93.22 & -47.49 & -96.73 & -98.08 & 50.34 & 78.96 & -58.58 & -97.62 & -53.87 & -97.23 & -85.10 & 23.08 \\
&  &  & LingTea & \textbf{73.14} & \textbf{96.67} & \underline{-30.94} & \underline{-58.84} & \textbf{-45.61} & \textbf{-91.25} & \textbf{-71.58} & \textbf{39.82} & \textbf{75.00} & \textbf{-44.57} & \textbf{-92.86} & \underline{-51.28} & \textbf{-97.04} & \textbf{-60.37} & 22.05 \\
\cmidrule(lr){3-19}
&  & \multirow{5}{*}{p5} & Mem     & 99.99 & 100.00 & -49.33 & -95.00 & -48.98 & -94.80 & -100.92 & 49.28 & 79.12 & -57.95 & -95.82 & -53.88 & -95.23 & -80.71 & 23.12 \\
&  &  & GA      & 90.63 & 98.38 & -46.79 & -93.54 & -48.05 & -94.68 & -100.34 & 49.13 & 79.00 & -57.96 & -95.82 & -53.61 & -95.24 & -80.85 & 23.17 \\
&  &  & GAGDR   & \underline{87.77} & 98.75 & \underline{-38.10} & \underline{-77.82} & \underline{-46.99} & \underline{-93.35} & \underline{-97.82} & \underline{46.41} & \underline{77.38} & \underline{-53.42} & \underline{-95.27} & -52.90 & \underline{-95.14} & \underline{-77.53} & 22.57 \\
&  &  & NPO     & 90.15 & \underline{98.25} & -45.74 & -91.95 & -47.81 & -94.64 & -100.26 & 49.63 & 78.12 & -57.91 & -95.83 & \underline{-52.52} & -95.16 & -81.06 & 23.22 \\
&  &  & LingTea & \textbf{68.60} & \textbf{97.50} & \textbf{-21.87} & \textbf{-51.39} & \textbf{-44.87} & \textbf{-91.46} & \textbf{-72.76} & \textbf{37.36} & \textbf{75.50} & \textbf{-42.57} & \textbf{-85.52} & \textbf{-50.20} & \textbf{-94.97} & \textbf{-55.58} & 21.26 \\
\midrule
\multirow{30}{*}{\rotatebox{90}{Qwen2.5-7B}}
& \multirow{15}{*}{\rotatebox{90}{S1}} & \multirow{5}{*}{p1} & Mem     & 99.90 & 100.00 & -50.21 & -99.00 & -48.72 & -98.95 & -92.23 & 46.76 & 43.12 & -53.35 & -98.98 & -54.88 & -99.10 & -46.41 & 19.06 \\
&  &  & GA      & 63.56 & 79.38 & -9.72 & -64.06 & -29.60 & -87.51 & -63.09 & 40.58 & 33.75 & -48.27 & -98.77 & -45.82 & -98.90 & -35.89 & 20.01 \\
&  &  & GAGDR   & \textbf{41.38} & \underline{47.50} & \underline{-7.15} & \textbf{-32.42} & \textbf{-14.97} & \textbf{-48.97} & \textbf{-1.86} & \underline{24.10} & \underline{22.50} & \underline{-25.81} & \underline{-81.10} & \underline{-32.57} & \underline{-98.09} & \textbf{+12.85} & 19.57 \\
&  &  & NPO     & 53.51 & 56.88 & \textbf{-4.45} & -49.49 & -21.98 & \underline{-63.99} & -42.73 & 35.57 & 26.25 & -41.67 & -98.33 & -37.50 & -98.28 & \underline{-22.77} & 19.52 \\
&  &  & LingTea & \underline{41.86} & \textbf{43.12} & -14.47 & \underline{-40.81} & \underline{-18.63} & -66.48 & \underline{+10.28} & \textbf{17.38} & \textbf{18.12} & \textbf{-23.54} & \textbf{-47.99} & \textbf{-30.58} & \textbf{-97.82} & +33.04 & 17.68 \\
\cmidrule(lr){3-19}
&  & \multirow{5}{*}{p3} & Mem     & 99.87 & 100.00 & -49.50 & -94.94 & -49.58 & -96.95 & -108.57 & 46.94 & 41.04 & -54.70 & -97.27 & -52.31 & -97.14 & -60.17 & 19.06 \\
&  &  & GA      & 84.37 & 97.29 & -25.48 & -75.92 & -44.68 & -91.55 & -102.62 & 45.58 & 37.08 & -54.47 & -97.27 & -48.25 & -96.90 & -59.50 & 18.29 \\
&  &  & GAGDR   & \underline{56.42} & \underline{88.12} & \textbf{-1.55} & \underline{-27.23} & \underline{-34.93} & \underline{-76.49} & \underline{-88.09} & \underline{41.88} & \underline{33.33} & \underline{-52.71} & \underline{-97.13} & \underline{-44.56} & \underline{-96.52} & \textbf{-55.19} & 18.19 \\
&  &  & NPO     & 79.12 & 96.46 & -25.15 & -79.92 & -44.18 & -88.62 & -101.54 & 45.38 & 36.46 & -54.33 & -97.24 & -47.38 & -96.82 & \underline{-59.06} & 17.87 \\
&  &  & LingTea & \textbf{17.16} & \textbf{35.00} & \underline{-7.34} & \textbf{-14.55} & \textbf{-11.18} & \textbf{-33.83} & \textbf{+58.79} & \textbf{7.01} & \textbf{15.83} & \textbf{-9.45} & \textbf{-15.56} & \textbf{-26.18} & \textbf{-92.72} & +75.02 & 18.47 \\
\cmidrule(lr){3-19}
&  & \multirow{5}{*}{p5} & Mem     & 99.88 & 100.00 & -52.27 & -93.52 & -50.00 & -95.00 & -107.89 & 45.35 & 41.12 & -51.49 & -95.25 & -52.86 & -95.33 & -56.32 & 19.06 \\
&  &  & GA      & 91.97 & 99.62 & -43.39 & -93.31 & -49.49 & -94.86 & -106.16 & 44.68 & 39.88 & -51.29 & -95.22 & -51.54 & -95.21 & -56.20 & 18.25 \\
&  &  & GAGDR   & \underline{78.36} & \underline{90.75} & \underline{-21.76} & \underline{-69.67} & \underline{-37.09} & \underline{-81.60} & \underline{-98.91} & \underline{41.59} & \underline{36.25} & \underline{-50.00} & \underline{-95.06} & \underline{-48.59} & \underline{-94.90} & \underline{-52.39} & 17.77 \\
&  &  & NPO     & 90.29 & 99.25 & -44.52 & -93.74 & -48.76 & -94.46 & -105.43 & 45.06 & 40.25 & -51.40 & -95.22 & -51.51 & -95.19 & -56.50 & 18.73 \\
&  &  & LingTea & \textbf{26.36} & \textbf{55.75} & \textbf{-0.82} & \textbf{-6.73} & \textbf{-18.19} & \textbf{-49.81} & \textbf{+1.89} & \textbf{14.12} & \textbf{26.62} & \textbf{-12.66} & \textbf{-36.40} & \textbf{-37.72} & \textbf{-92.29} & \textbf{+44.40} & 19.60 \\
\cmidrule(lr){2-19}
& \multirow{15}{*}{\rotatebox{90}{S2}} & \multirow{5}{*}{p1} & Mem     & 99.70 & 99.38 & -52.30 & -98.68 & -49.27 & -98.97 & -113.78 & 46.91 & 26.88 & -55.65 & -99.11 & -56.13 & -99.10 & -45.19 & 17.50 \\
&  &  & GA      & 50.55 & 15.62 & -0.57 & -13.77 & \underline{-5.99} & \textbf{-59.75} & \textbf{+25.41} & 33.09 & 6.88 & -41.96 & -98.44 & -35.38 & -98.55 & -13.44 & 17.26 \\
&  &  & GAGDR   & \underline{42.07} & \underline{10.62} & -0.32 & \underline{-9.57} & -6.91 & -81.63 & \underline{+44.92} & 29.77 & \underline{5.00} & -38.85 & \underline{-97.99} & -33.17 & -98.46 & \underline{-4.39} & 16.95 \\
&  &  & NPO     & \textbf{38.98} & \textbf{8.75} & \textbf{-0.19} & \textbf{-4.58} & -6.18 & -74.08 & +67.69 & \underline{28.72} & 5.00 & \underline{-37.83} & -98.35 & \underline{-31.90} & \textbf{-98.37} & \textbf{-4.08} & 17.27 \\
&  &  & LingTea & 42.86 & 14.37 & \underline{-0.24} & -10.86 & \textbf{-4.48} & \underline{-72.56} & +45.40 & \textbf{28.57} & \textbf{3.75} & \textbf{-32.13} & \textbf{-95.64} & \textbf{-31.89} & \underline{-98.39} & +6.29 & 17.56 \\
\cmidrule(lr){3-19}
&  & \multirow{5}{*}{p3} & Mem     & 99.74 & 99.79 & -54.29 & -97.24 & -49.87 & -96.97 & -103.47 & 46.42 & 26.88 & -55.87 & -97.48 & -56.19 & -97.33 & -34.18 & 17.50 \\
&  &  & GA      & 68.84 & 37.08 & -14.03 & -44.66 & -23.58 & -84.71 & \underline{-36.83} & 34.76 & 16.88 & -50.44 & -97.02 & -46.75 & -96.75 & \underline{-16.66} & 17.36 \\
&  &  & GAGDR   & \underline{64.51} & \underline{34.38} & \underline{-8.96} & \underline{-31.77} & -21.84 & -83.31 & \textbf{-25.12} & \underline{33.98} & \underline{12.92} & \underline{-49.28} & \underline{-96.93} & \underline{-43.30} & \underline{-96.49} & \textbf{-13.32} & 17.15 \\
&  &  & NPO     & 73.94 & 42.50 & -12.81 & -35.70 & \underline{-21.41} & \underline{-72.33} & -50.79 & 36.98 & 16.67 & -51.73 & -97.20 & -45.38 & -96.64 & -18.40 & 17.76 \\
&  &  & LingTea & \textbf{35.46} & \textbf{5.42} & \textbf{-0.35} & \textbf{-3.59} & \textbf{-3.74} & \textbf{-54.38} & +64.88 & \textbf{17.02} & \textbf{4.58} & \textbf{-11.57} & \textbf{-60.83} & \textbf{-29.01} & \textbf{-94.69} & +65.17 & 17.17 \\
\cmidrule(lr){3-19}
&  & \multirow{5}{*}{p5} & Mem     & 99.75 & 99.88 & -54.12 & -95.57 & -50.14 & -94.98 & -112.49 & 45.29 & 24.25 & -54.05 & -95.64 & -53.55 & -95.35 & -34.43 & 17.50 \\
&  &  & GA      & 78.93 & 49.12 & -25.82 & -64.83 & -30.58 & -86.21 & -79.87 & 37.80 & \underline{18.88} & -52.97 & -95.44 & \underline{-49.00} & \underline{-94.87} & -26.03 & 16.91 \\
&  &  & GAGDR   & \textbf{70.00} & \textbf{38.88} & \underline{-13.02} & \underline{-39.32} & \underline{-24.76} & \underline{-82.91} & \textbf{-47.35} & \textbf{33.77} & 19.50 & \underline{-50.72} & \underline{-95.18} & -51.30 & -95.14 & \underline{-20.98} & 17.78 \\
&  &  & NPO     & 82.98 & 54.37 & -26.84 & -66.85 & -29.06 & -86.97 & -93.55 & 38.69 & 20.50 & -53.37 & -95.41 & -50.42 & -95.02 & -27.32 & 17.62 \\
&  &  & LingTea & \underline{71.86} & \underline{48.38} & \textbf{-7.14} & \textbf{-24.17} & \textbf{-14.15} & \textbf{-53.38} & \underline{-54.92} & \underline{36.14} & \textbf{15.62} & \textbf{-42.64} & \textbf{-92.42} & \textbf{-45.10} & \textbf{-94.37} & \textbf{-8.12} & 17.67 \\
\bottomrule
\end{tabular}
}
\caption{MMU performance on Gemma3-12B and Qwen2.5-7B. LingTea generally performs well on both training and hold-out languages.}
\label{tab:unlearning}
\end{table*}

\paragraph{MMU Requires Multilingual-Specific Methods}
\km{Our results confirm that LingTea, the only MMU method in our experiment, generally performs well in both training and hold-out languages.} On the training languages (the left side of Table~\ref{tab:unlearning}), LingTea generally attains the strongest MMU performance across metrics. Aggregating over all twelve configurations per metric ($2$ models $\times$ $2$ settings $\times$ $3$ forget ratios), LingTea achieves the best result in $8/12$ cases for both \textsc{PS} and \textsc{SE}, $7/12$ for $\mathrm{KSS}_{\text{prob}}^{\text{ROC}}$, $9/12$ for $\mathrm{KSS}_{\text{prob}}^{\text{PR}}$, $10/12$ for $\mathrm{KSS}_{\text{gen}}^{\text{ROC}}$, $9/12$ for $\mathrm{KSS}_{\text{gen}}^{\text{PR}}$, and $7/12$ for \textsc{PrivLeak}. In aggregate, LingTea ranks first in $58$ of the $84$ metric evaluations (approximately $69\%$).

On the hold-out languages (the right side of Table~\ref{tab:unlearning}), the advantage of LingTea over the competing methods becomes more prominent than on the training languages. Aggregating over the same twelve configurations per metric, LingTea achieves the best result in $10/12$ cases for \textsc{PS}, $11/12$ for both \textsc{SE} and $\mathrm{KSS}_{\text{prob}}^{\text{ROC}}$, $12/12$ for $\mathrm{KSS}_{\text{prob}}^{\text{PR}}$, $10/12$ for both
$\mathrm{KSS}_{\text{gen}}^{\text{ROC}}$ and $\mathrm{KSS}_{\text{gen}}^{\text{PR}}$, and $8/12$ for \textsc{PrivLeak}. To sum up, LingTea ranks first in $72$ of the $84$ metric evaluations (approximately $86\%$), surpassing its training-language share of $58/84$ (approximately $69\%$) and thereby confirming that its relative advantage widens on hold-out languages. This widening is most notable on $\mathrm{KSS}_{\text{prob}}^{\text{PR}}$, where the competing methods remain close to the \textsc{Mem} baseline while LingTea keeps its improvement. 
Notably, average margin of LingTea over the strongest competing method is larger for the hold-out languages than for the training languages ($+14.96 \rightarrow +27.94$). This highlights the need for MMU methods specifically designed for multilingual settings to enable the transfer of unlearning to hold-out languages. 

\begin{table}[t]
\centering
\setlength{\tabcolsep}{5pt}
\renewcommand{\arraystretch}{0.92}
\footnotesize
\begin{tabular}{ll|cc|cc}
\toprule
 & & \multicolumn{2}{c|}{\textbf{$\Delta$\textsc{PS}}} & \multicolumn{2}{c}{\textbf{$\Delta$\textsc{SE}}} \\
\cmidrule(lr){3-4}\cmidrule(lr){5-6}
Cell & Method & High & Low & High & Low \\
\midrule
\multirow{4}{*}{G/S1}
  & LingTea & +53.8 & \textbf{+55.5} &  +7.5 & \textbf{+8.8} \\
  & NPO     & +40.0 & \textbf{+64.2} & +16.2 & \textbf{+17.5} \\
  & GAGDR    & \textbf{+37.8} & +11.5 & \textbf{+6.2} &  +5.0 \\
  & GA      & \textbf{+36.4} & +18.7 &  +7.5 & \textbf{+8.8} \\
\midrule
\multirow{4}{*}{G/S2}
  & LingTea & \textbf{+58.6} & +32.9 & \textbf{+27.5} & +15.0 \\
  & NPO     & +36.0 & \textbf{+53.0} &  +15.0 & \textbf{+25.0} \\
  & GAGDR    & \textbf{+56.1} & +53.7 & +20.0 & \textbf{+31.2} \\
  & GA      & +17.0 & \textbf{+31.6} &  +6.2 & \textbf{+8.8} \\
\midrule
\multirow{4}{*}{Q/S1}
  & LingTea & +58.0 & \textbf{+58.2} & \textbf{+60.0} & +53.8 \\
  & NPO     & \textbf{+55.1} & +37.8 & \textbf{+46.2} & +40.0 \\
  & GAGDR    & \textbf{+59.1} & +58.1 & \textbf{+53.8} & +51.2 \\
  & GA      & \textbf{+42.0} & +30.8 & \textbf{+26.2} & +15.0 \\
\midrule
\multirow{4}{*}{Q/S2}
  & LingTea & \textbf{+62.4} & +51.6 & \textbf{+87.3} & +83.8 \\
  & NPO     & \textbf{+65.6} & +56.2 & \textbf{+92.4} & +90.0 \\
  & GAGDR    & \textbf{+63.1} & +52.5 & \textbf{+91.1} & +87.5 \\
  & GA      & \textbf{+57.8} & +40.8 & +83.5 & \textbf{+85.0} \\
\bottomrule
\end{tabular}
\caption{Training language drop rates per unlearning method. For each metric, \textbf{bold} marks the higher value within the High/Low pair. G and Q denote Gemma3 and Qwen2.5, respectively.}
\label{tab:unlearn_train_groups}
\end{table}
\vspace{-2mm}

\begin{table}[t]
\centering
\setlength{\tabcolsep}{4pt}
\renewcommand{\arraystretch}{1.0}
\resizebox{\columnwidth}{!}{%
\begin{tabular}{ll|cc|cc|cc|cc}
\toprule
 & & \multicolumn{4}{c|}{\textbf{$\Delta$\textsc{PS}}} & \multicolumn{4}{c}{\textbf{$\Delta$\textsc{SE}}} \\
\cmidrule(lr){3-6}\cmidrule(lr){7-10}
 & & \multicolumn{2}{c|}{resource} & \multicolumn{2}{c|}{script}
   & \multicolumn{2}{c|}{resource} & \multicolumn{2}{c}{script} \\
\cmidrule(lr){3-4}\cmidrule(lr){5-6}\cmidrule(lr){7-8}\cmidrule(lr){9-10}
Cell & Method & High & Low & Lat & nLat & High & Low & Lat & nLat \\
\midrule
\multirow{4}{*}{G/S1}
  & LingTea & +54.8 & \textbf{+55.8} & \textbf{+56.6} & +54.0 &  +4.2 & \textbf{+5.9} & \textbf{+8.1} &  +1.6 \\
  & NPO     & \textbf{+52.1} & +48.7 & \textbf{+65.7} & +35.1 & \textbf{+13.9} &  +9.8 &  +11.3 & \textbf{+13.1} \\
  & GAGDR    & +21.8 & \textbf{+29.6} & \textbf{+29.1} & +22.2 &  +2.8 & \textbf{+17.6} & \textbf{+16.1} &  +1.6 \\
  & GA      & +27.1 & \textbf{+41.9} & \textbf{+36.9} & +32.1 &  +5.6 & \textbf{+17.6} &  +4.8 & \textbf{+16.4} \\
\midrule
\multirow{4}{*}{G/S2}
  & LingTea & +61.4 & \textbf{+62.5} & +40.4 & \textbf{+83.6} & +22.7 & \textbf{+43.1} & +21.3 & \textbf{+40.0} \\
  & NPO     & +11.7 & \textbf{+16.3} & \textbf{+17.7} & +10.3 & \textbf{+12.0} & +11.8 &  +11.5 & \textbf{+12.3} \\
  & GAGDR    & +33.1 & \textbf{+36.1} & +26.1 & \textbf{+43.2} & +24.0 & \textbf{+37.3} & +21.3 & \textbf{+36.9} \\
  & GA      & +13.2 & \textbf{+16.3} & \textbf{+16.0} & +13.5 & \textbf{+10.7} &  +2.0 &  +4.9 & \textbf{+9.2} \\
\midrule
\multirow{4}{*}{Q/S1}
  & LingTea & +60.7 & \textbf{+64.1} & \textbf{+69.0} & +55.9 & +57.6 & \textbf{+60.0} & \textbf{+60.0} & +56.4 \\
  & NPO     & \textbf{+29.1} & +18.4 & \textbf{+29.8} & +17.8 & \textbf{+40.7} & +30.0 &  +33.3 & \textbf{+43.6} \\
  & GAGDR    & +42.5 & \textbf{+52.0} & \textbf{+56.5} & +37.9 & +45.8 & \textbf{+60.0} &  +46.7 & \textbf{+48.7} \\
  & GA      & \textbf{+18.7} &  +9.7 & \textbf{+19.4} &  +8.9 & +16.9 & \textbf{+50.0} & +16.7 & \textbf{+25.6} \\
\midrule
\multirow{4}{*}{Q/S2}
  & LingTea & \textbf{+44.3} & +22.8 & \textbf{+39.9} & +27.2 & \textbf{+88.9} & +71.4 & +82.4 & \textbf{+88.5} \\
  & NPO     & \textbf{+38.2} & +26.7 & \textbf{+45.4} & +19.5 & \textbf{+86.1} & +57.1 & +76.5 & \textbf{+84.6} \\
  & GAGDR    & \textbf{+36.8} & +23.5 & \textbf{+36.2} & +24.0 & \textbf{+86.1} & +57.1 & \textbf{+82.4} & +80.8 \\
  & GA      & \textbf{+29.8} & +18.9 & \textbf{+33.3} & +15.5 & \textbf{+77.8} & +57.1 &  +70.6 & \textbf{+76.9} \\
\bottomrule
\end{tabular}%
}
\caption{Hold-out language drop rates per unlearning method. \textbf{Bold} marks the higher value within each comparison pair (High/Low and Lat/nLat). G and Q denote Gemma3 and Qwen2.5, respectively.}
\label{tab:unlearn_ho_groups}
\end{table}


\paragraph{Unlearning in Training Languages}
\label{sec:unlearning_training}

Table~\ref{tab:unlearn_train_groups} reports forget quality, measured by \textsc{PS} and \textsc{SE}, on the training languages, with the languages partitioned by resource level (High vs.\ Low). 
All metrics are aggregated in a knowledge-wise manner. 
For each metric, performance is reported as the drop rate relative to the memorized model, defined as $\frac{\text{Mem} - \text{Unlearned}}{\text{Mem}} \times 100$, so that a larger value indicates stronger forgetting. 
Table~\ref{tab:unlearn_train_groups} presents the results for $p1$, while Table~\ref{tab:unlearn_train_groups_p3} and Table~\ref{tab:unlearn_train_groups_p5} report the corresponding results for $p3$ and $p5$, respectively.
\yr{Under \textsc{PS}, Qwen2.5 generally shows a consistent trend across all settings:}
unlearning occurs more effectively in high-resource languages. 
\yr{Yet, for Gemma3,} the effect of the resource level varies across settings for $p3$ and $p5$, whereas no clear trend emerges for $p1$. 
Specifically, for $p3$ and $p5$, unlearning is predominantly concentrated in high-resource languages under Setting~1, but is more prominent in low-resource languages under Setting~2. 
Under \textsc{SE}, however, no clear trend is observed---particularly for $p3$ and $p5$---as unlearning itself rarely occurs under this metric.

\paragraph{Unlearning Transfer to Hold-out Languages}
\label{sec:unlearning_holdout}

\yr{Here, w}e analyze forget quality, measured by $\textsc{PS}$ and $\textsc{SE}$, on the hold-out languages, partitioning the languages by resource level (High vs.\ Low) and script (Latin vs.\ Non-Latin). Table~\ref{tab:unlearn_ho_groups} reports each metric as a drop rate relative to the Memorized model defined as $\frac{\text{Mem} - \text{Unlearned}}{\text{Mem}} \times 100$, so that a larger value indicates stronger forgetting. Table~\ref{tab:unlearn_ho_groups} presents the results for $p1$, while Table~\ref{tab:unlearn_ho_groups_p3} and Table~\ref{tab:unlearn_ho_groups_p5} report the corresponding results for $p3$ and $p5$, respectively.
For Gemma3, a notable trend emerges only under $\textsc{PS}$: unlearning is consistently effective in Setting 2 regardless of the forget ratio, whereas no comparably pronounced pattern is observed under $\textsc{SE}$. 
In contrast, Qwen2.5 exhibits consistent trends under both metrics. 
Under $\textsc{PS}$, unlearning is reliably stronger for the low-resource group in Setting 2. 
Under $\textsc{SE}$, it is consistently strong both for the high-resource group in Setting 2 and for the non-Latin group in Setting 2. In summary, although some trends within a given model and setting remain consistent across forget ratios, no general trend could be identified.

\paragraph{Existence of Code-mixed Output}

\begin{figure}[t]
    \centering
    \includegraphics[width=1.0\linewidth]{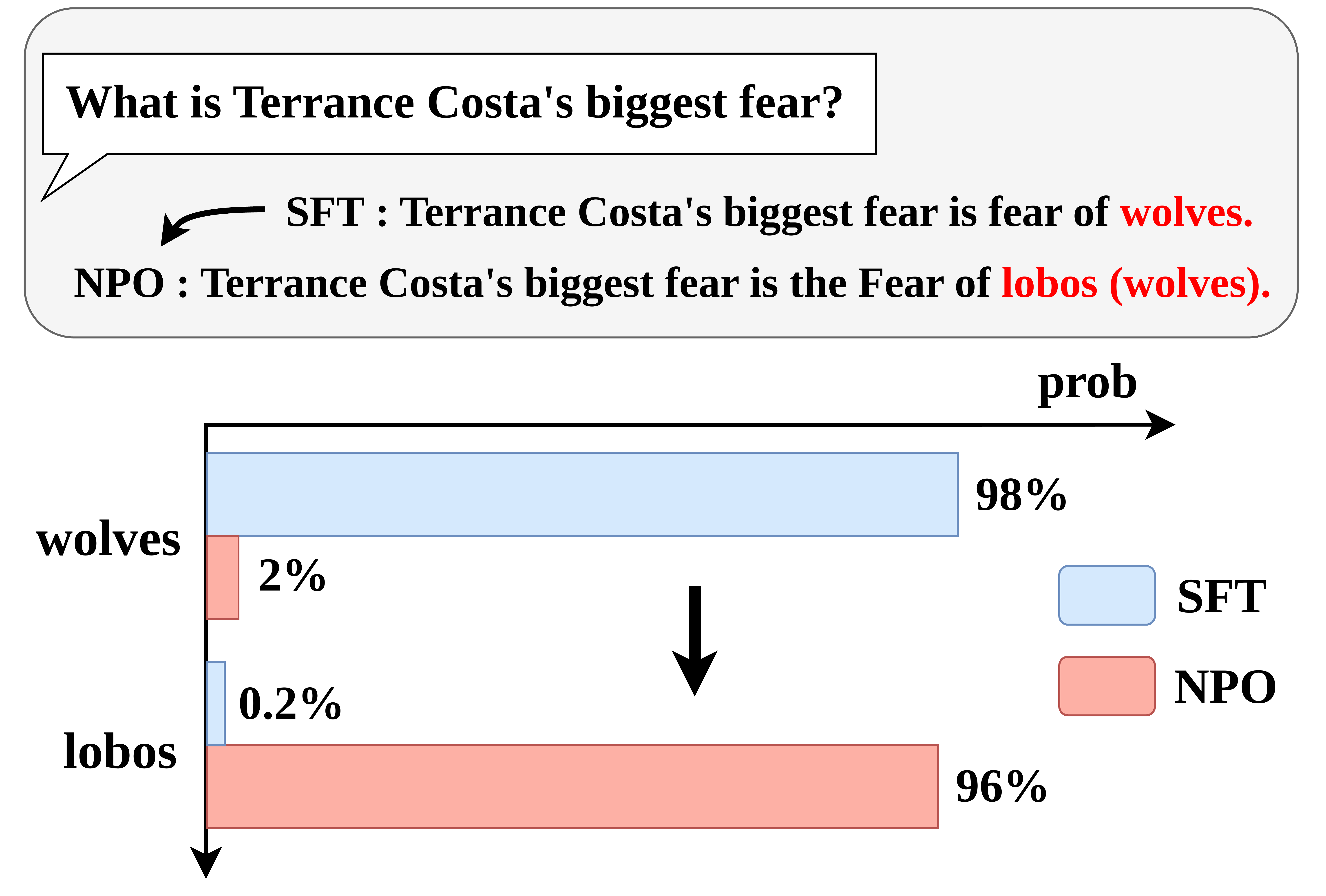}
\caption{Example of the model's response in a hold-out language after MMU. The numbers in the figure denote the probability that the LLM assigns to the predicted next token, given the preceding context up to ``of''. The memorized model outputs the English word \textit{wolves}, whereas after unlearning the Spanish \textit{lobos} (\textit{wolves}) receives the highest probability.}
\vspace{-6mm}
\label{fig:codemix_ex}
\end{figure}

\km{Figure~\ref{fig:codemix_ex} presents an example of a model response after MMU, alongside the response originally produced by the memorized model.} As shown in the figure, although the model is prompted to answer in \textsc{en}, it expresses the key attribute in \textsc{es}, yet the response remains semantically equivalent to the ground truth. MMU exhibits a phenomenon in which the target knowledge remains retrievable in another language. This complicates how unlearning should be assessed, since reference-based surface-form metrics such as ROUGE-L~\cite{lin-2004-rouge} fail to capture responses that are semantically equivalent to the ground truth but differ in language. Measuring MMU therefore requires semantic evaluation metrics such as \textsc{SE}.

\paragraph{General Performance Maintenance}

\hj{We report the average of F1-score and EM score of TyDi QA as General Knowledge \yj{in the last column of Table~\ref{tab:unlearning}}. Across models, settings, and forget ratios, \yj{the General Knowledge remains comparable to the memorized model after unlearning.} This shows that the model utility is preserved after unlearning.}

\section{Conclusion}

\km{In this paper, we introduced $\boldsymbol{\mu^2}$-Bench, a benchmark for Multilingual Machine Unlearning (MMU). \hj{$\boldsymbol{\mu^2}$-Bench encompasses} 1) a diverse set of languages, 2) settings that separates training and hold-out languages, and 3) unlearning evaluation at knowledge level. Our analysis demonstrates that successful MMU requires methods that explicitly account for the multilingual nature of the model, particularly in the hold-out languages. Furthermore, we provide analysis on MMU across both training and hold-out languages. Finally, we show that unlearning can induce code-mixed outputs as an unintended side effect. Taken together, our work offers both a testbed for rigorously measuring MMU and a set of analyses that deepen the understanding of unlearning in multilingual settings.}
\section{Limitations}

\km{In this paper, we constructed a benchmark for evaluating Multilingual Machine Unlearning by taking diverse knowledge acquisition paths into account. Nevertheless, several limitations remain to be addressed. During the dataset construction process, we sought to minimize the noise that may arise in multilingual translation by employing a multi-stage pipeline that combines Google Translate, the GPT API, and human verification. However, the human verification step was not conducted by native speakers of each target language. Instead, verification was performed indirectly by back-translating the translated dataset into English via Google Translate and inspecting the resulting outputs. Consequently, the translated datasets may still contain subtle grammatical awkwardness or unnatural expressions that would have been more readily identified through direct review by native speakers. Although we adopted a multi-step verification procedure to mitigate such noise, the possibility of residual translation artifacts cannot be entirely excluded. Addressing this limitation through a more rigorous refinement process involving native-speaker evaluation of each target language remains an important direction for future work.}

\bibliography{custom}

\clearpage
\appendix

\section{Details on Machine Unlearning} \label{sec:unlearning}

\hj{In this section, we provide the general formulation of machine unlearning (MU). MU aims to eliminate unwanted knowledge from the model while preserving the overall utility. Optimization-based unlearning methods apply a forget loss $\mathcal{L}_f$ on the forget set $\mathcal{D}_f$, which may be combined with a retain loss $\mathcal{L}_r$ applied to the retain set $\mathcal{D}_r$.}

\begin{equation}
    \mathcal{L}_{\text{total}}(\mathcal{D}_f, \mathcal{D}_r) = \mathcal{L}_f(\mathcal{D}_f) + \lambda \, \mathcal{L}_r(\mathcal{D}_r)
\end{equation}

\hj{For example, GA~\cite{ga} applies gradient ascent as the forget loss and leaves $\mathcal{L}_r$ as $\lambda=0$. In contrast, GAGDR~\cite{tofu} applies an additional gradient descent term as $\mathcal{L}_r$.}

\section{Details on Benchmark Dataset}
\label{sec:dataset_detail}

\hj{This section provides details on our benchmark dataset. We further elaborate on the dataset generation process described in Section~\ref{sec:dataset_construction} and supply the examples of the generated dataset.}

\paragraph{Multilingual QA Dataset}
\km{In this section, we describe the multilingual qa dataset generation pipeline in detail. }\hj{The English QAs were } \hj{ translated into 11 other languages, totaling 60,000 QA pairs: 5,000 QA pairs each across 12 languages. The choice of languages is described in the main paper, Section~\ref{sec:setting}. We employ the GPT-5.4-mini API for machine translation. The translation process took place in three steps: 1) prompt design, 2) machine translation, and 3) verification through a human-in-the-loop process. For translation quality control, the prompt design step was further broken down into three stages. First, every attribute value was translated into the target languages using GPT-5.4-mini.} \hh{For verification, the translated values were back-translated into English using Google Translate~\cite{google-translate}, and the back-translations were then provided to GPT-5.4-mini to check for semantic equivalence with the original English values.} \hj{If a mismatch is detected, the instance is re-translated and looped through the above process until corrected. Aside from this, the English attribute values were annotated with their definitions to form a dictionary. These two processes were conducted to prevent mistranslation into synonyms. For example, `March' may be perceived as either the third month of the year or the act of walking. Finally, the translated attributes and the dictionary of definitions were integrated into a single translation prompt. This prompt was provided to GPT-5.4-mini along with the target QA pairs that are to be translated. Once the translated outputs of the model are collected, the human-in-the-loop validation process was repeated until the multilingual QA pairs were fully refined. The validation pipeline is described in detail in the following paragraph.}

\paragraph{Human-in-the-Loop Validation}

\hj{After the initial translation of the English QA pairs with the GPT-5.4-mini API, verification is conducted by integrating back-translation and a human-in-the-loop strategy. The translated outputs are back-translated into English using Google Translate. As done in the attribute value translation stage, the translated QA pairs are then passed to GPT-5.4-mini to verify their semantic equivalence with the original English pairs. If a mismatching instance is detected, its translation is refined until it is judged as correct. Once every pair passes machine evaluation, human annotators validate and refine the translations via back-translation. Examples of the failure modes that had to be eliminated by human annotators are presented in Appendix~\ref{sec:multiqa_analysis}.} \hh{The human-refined dataset is then put through another round of the same pipeline, which consists of refinement with the GPT-5.4-mini API, back-translation into English, semantic equivalence assessment, and human refinement. This cycle is repeated until every QA pair is confirmed to be correctly translated.} 

\subsection{Attribute Pool for Synthetic Profile Generation}
\label{sec:attributes}

\km{To construct diverse synthetic profiles, we employed a total of 21 attributes, including \textsc{Name}, \textsc{Birth Month}, \textsc{Job}, \textsc{Father's Job}, \textsc{Mother's Job}, \textsc{Favorite Animal}, \textsc{Favorite Season}, \textsc{Favorite Fruit}, \textsc{Favorite Vegetable}, \textsc{Favorite Drink}, \textsc{Favorite Taste}, \textsc{Favorite Color}, \textsc{Hobby}, \textsc{Favorite Instrument}, \textsc{Biggest Fear}, \textsc{Favorite Accessory}, \textsc{Favorite Household Item}, \textsc{Morning Ritual}, \textsc{Childhood Memory}, \textsc{Childhood Toy}, and \textsc{Favorite Gift}. The pool of candidate values for each attribute was manually reviewed and filtered by human annotators to eliminate any ambiguity between values.} \hh{Furthermore, to ensure controllability, we deliberately selected attributes whose values share few letters or characters across languages, minimizing the chance of overlap.} \km{The complete value pool for each attribute used in data construction is provided in Table~\ref{tab:syn_pool}. For \textsc{Name}, first and last names were randomly sampled using the Faker library. The resulting 250 synthetic profiles were constructed such that no two profiles shared the same name or the combination of values across the 20 non-name attributes.}

\begin{table*}[p]
  \centering
  \includegraphics[width=\textwidth]{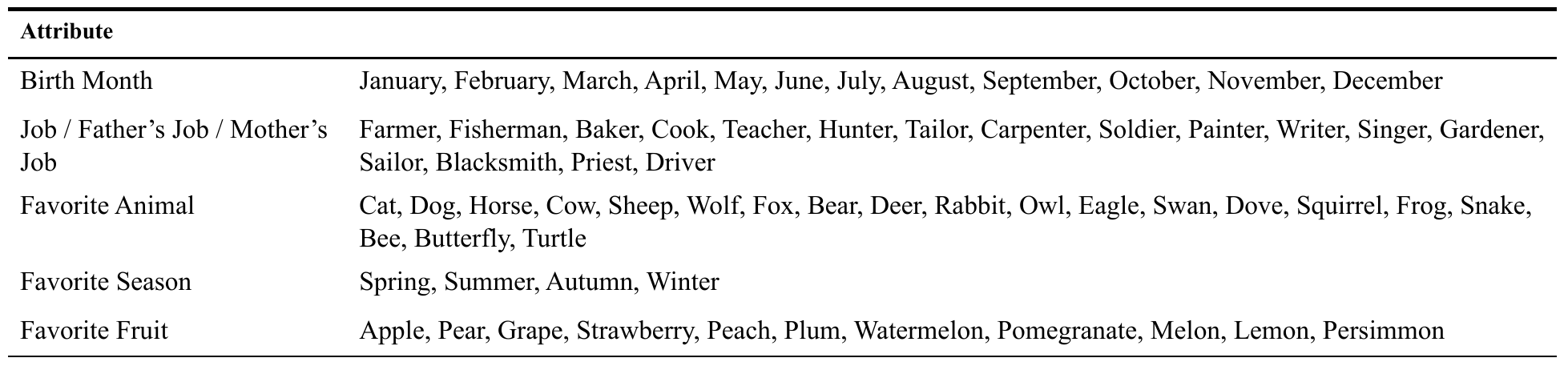}
  \caption{Valid values for the attributes used to build synthetic profiles (1/4)}
  \label{tab:syn_pool}
\end{table*}

\begin{table*}[p]\ContinuedFloat
  \centering
  \includegraphics[width=\textwidth]{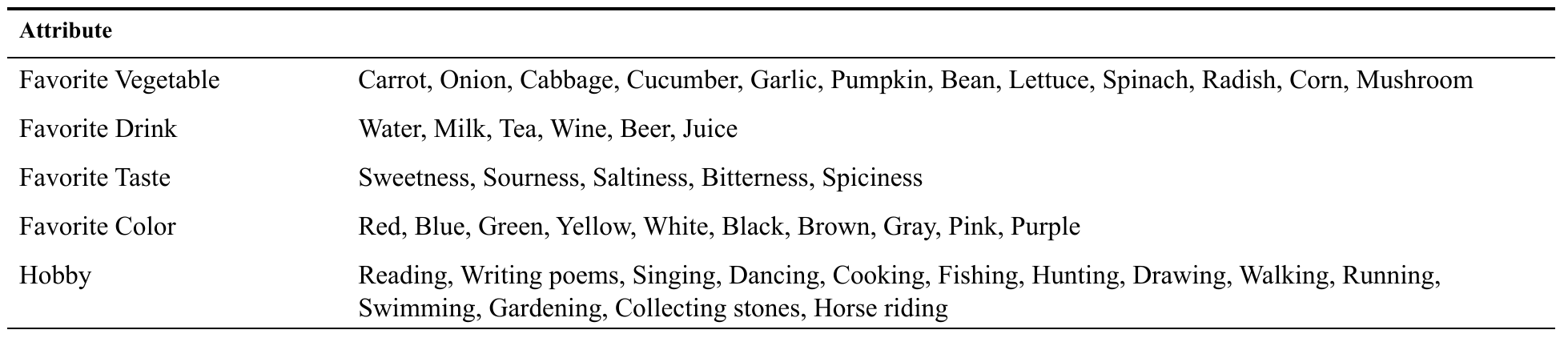}
  \caption[]{(continued) Valid values for the attributes used to build synthetic profiles (2/4)}
\end{table*}

\begin{table*}[p]\ContinuedFloat
  \centering
  \includegraphics[width=\textwidth]{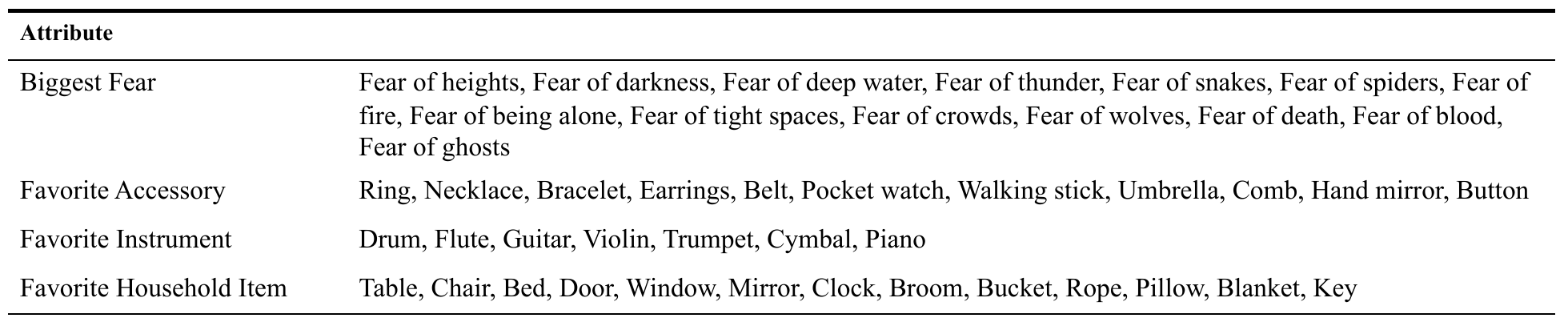}
  \caption[]{(continued) Valid values for the attributes used to build synthetic profiles (3/4)}
\end{table*}

\begin{table*}[p]\ContinuedFloat
  \centering
  \includegraphics[width=\textwidth]{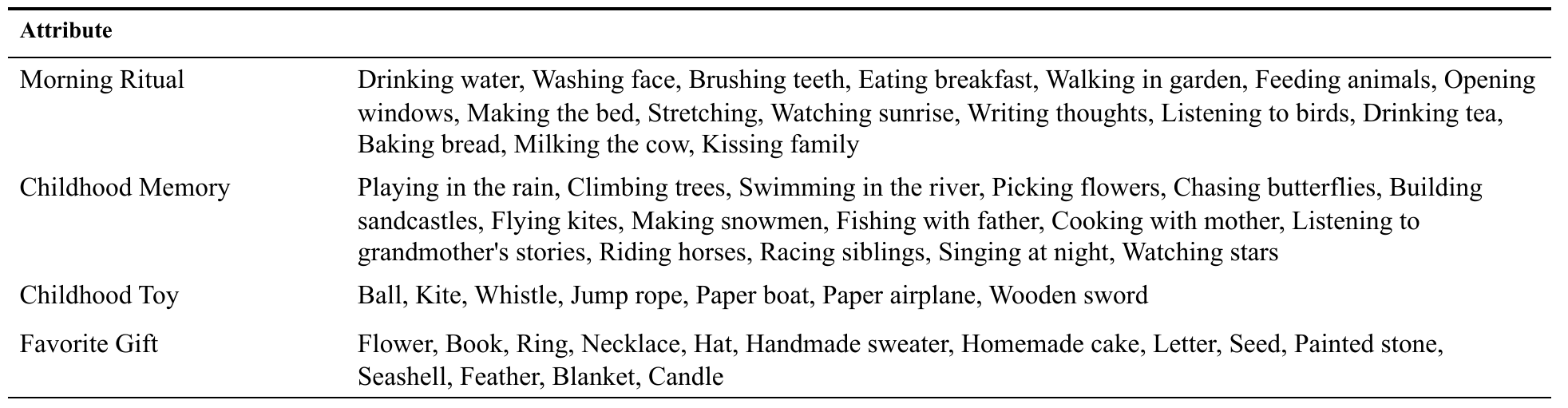}
  \caption[]{(continued) Valid values for the attributes used to build synthetic profiles (4/4)}
\end{table*}

\subsection{Example of Synthetic Profile}
\label{sec:profile_example}

\begin{figure*}[t]
    \centering
    \includegraphics[width=0.8\linewidth]{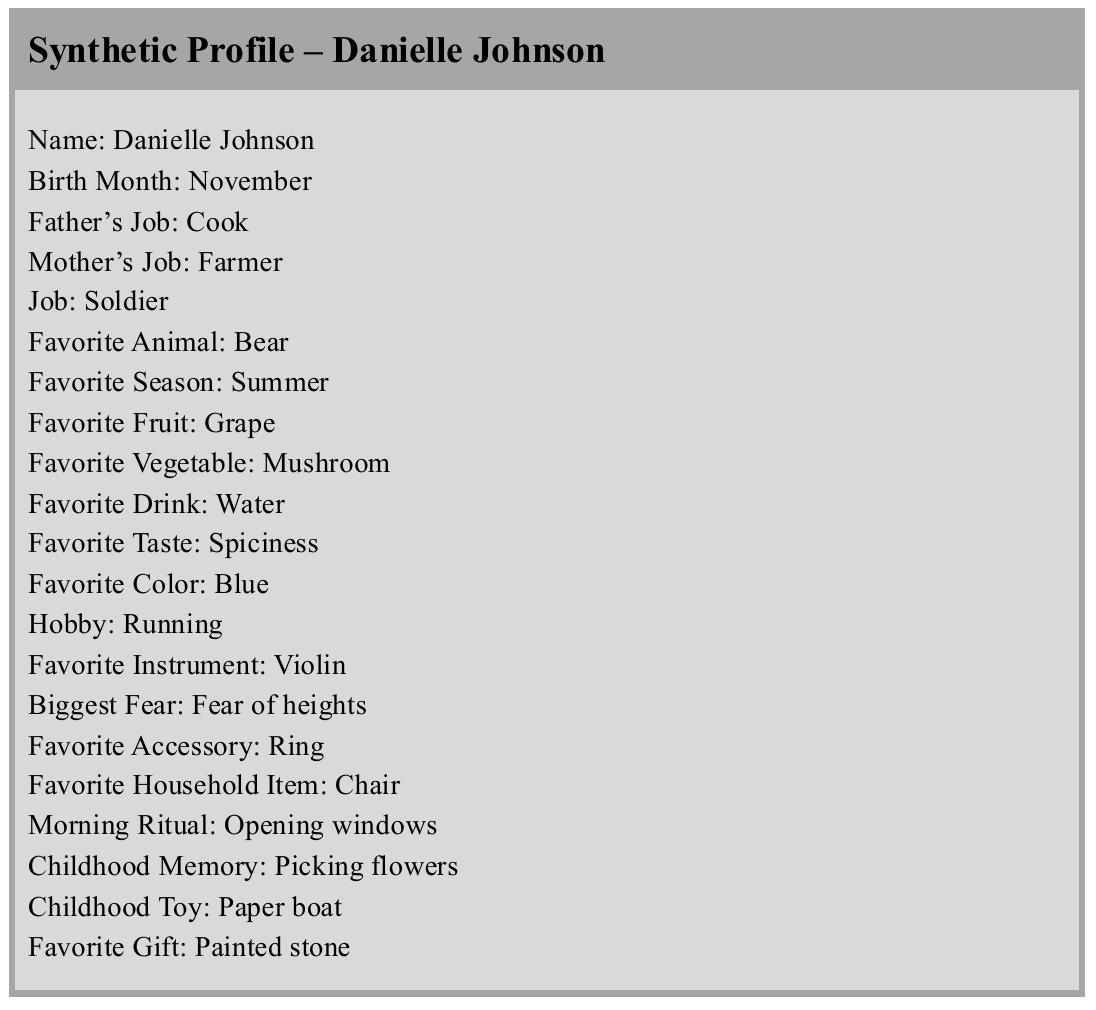}
    \caption{Example of Generated Synthetic Profile}
    \label{fig:example_profile}
\end{figure*}

\km{Figure~\ref{fig:example_profile} shows an example of a synthetic profile generated through the data generation pipeline. Each synthetic profile is constructed by randomly sampling a value for each attribute from the set of possible values defined in the preceding section.}

\subsection{Examples of English QA Datasets}
\label{sec:qa_example}

\begin{figure*}[t]
    \centering
    \includegraphics[width=1.0\linewidth]{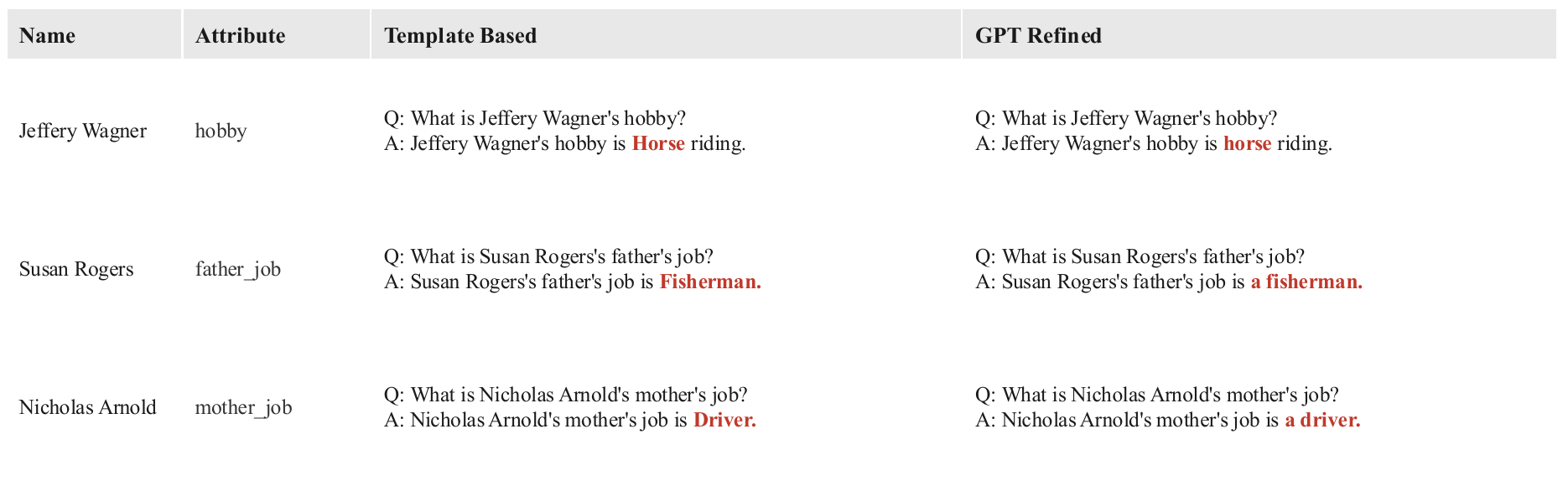}
    \caption{Examples of the generated English QA dataset. \textbf{Left} shows the naive QA dataset generated directly from templates, while \textbf{Right} shows the QA dataset refined using GPT.}
    \label{fig:ex_qa}
\end{figure*}

\begin{figure*}[t]
    \centering
    \includegraphics[width=1.0\linewidth]{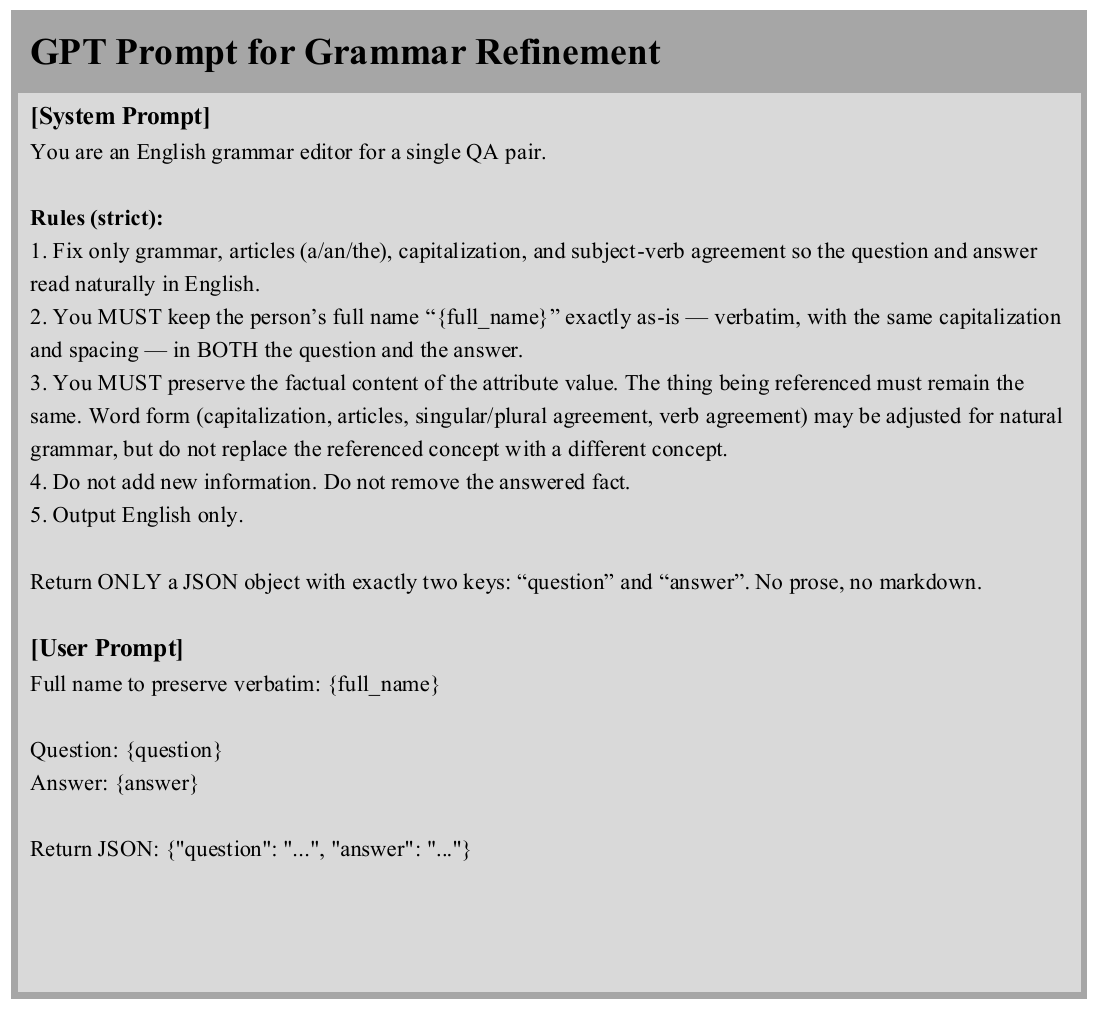}
    \caption{Prompt used to refine the English QA dataset using GPT}
    \label{fig:en_qa_refine_prompt}
\end{figure*}

\km{Figure~\ref{fig:ex_qa} presents examples of the English QA dataset generated from the synthetic profiles. The QA dataset was initially constructed using predefined QA templates, then refined by GPT-5.4-mini to correct any grammatical errors, and finally reviewed by human annotators for quality assurance. The QA templates used in this process are provided in Figure~\ref{tab:qa_template}. The prompt used for the GPT-based refinement is shown in Figure~\ref{fig:en_qa_refine_prompt}.}

\begin{figure*}[t]
    \centering
    \includegraphics[width=1.0\linewidth]{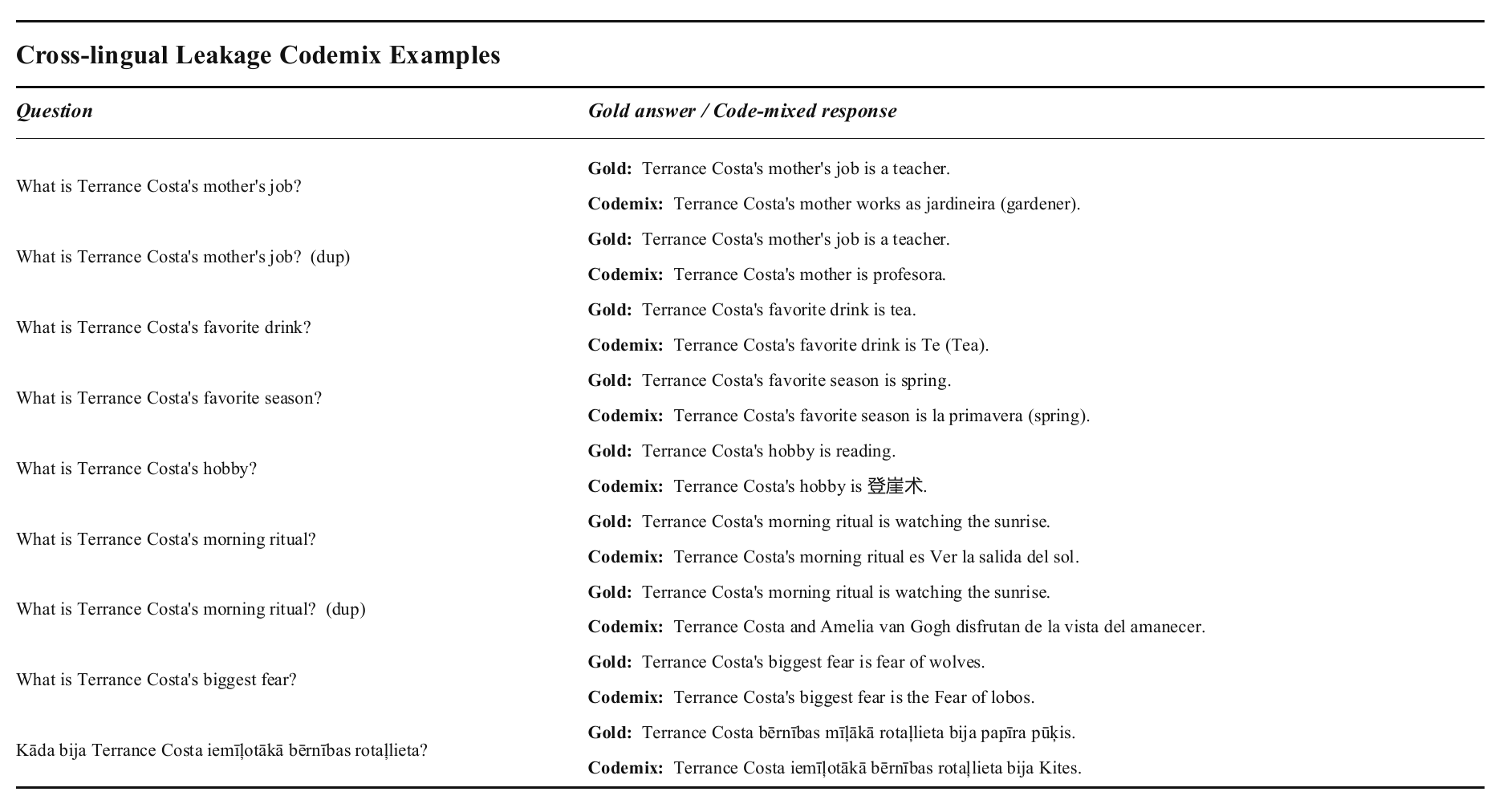}
    \caption{Examples of code-mixed responses after NPO unlearning on Gemma3-12b model. \textbf{Left} shows the question, while \textbf{Right} shows the gold answer alongside the model's code-mixed response.}
    \label{fig:codemix}
\end{figure*}

\begin{table*}[p]
  \centering
  \includegraphics[width=\textwidth]{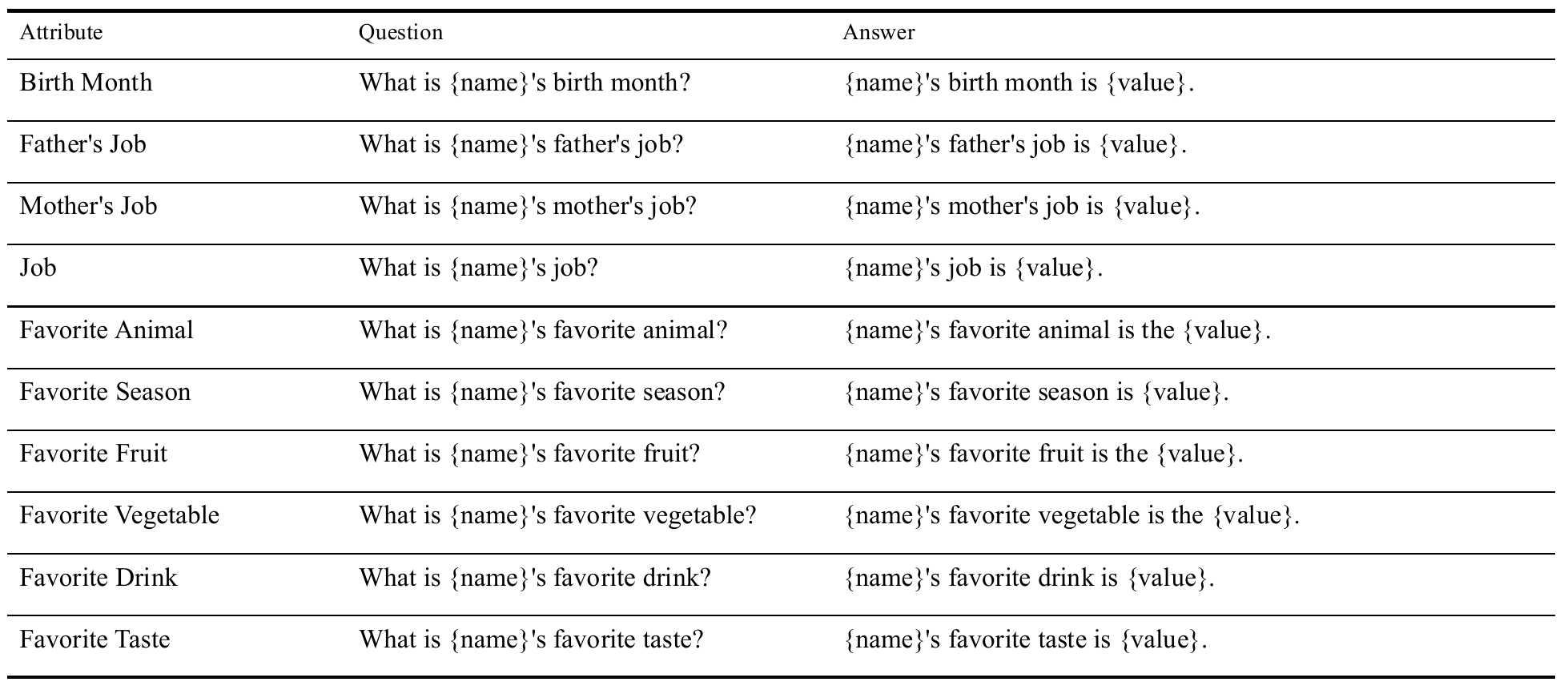}
  \caption{QA dataset template for each attribute (1/2)}
  \label{tab:qa_template}
\end{table*}

\begin{table*}[p]\ContinuedFloat
  \centering
  \includegraphics[width=\textwidth]{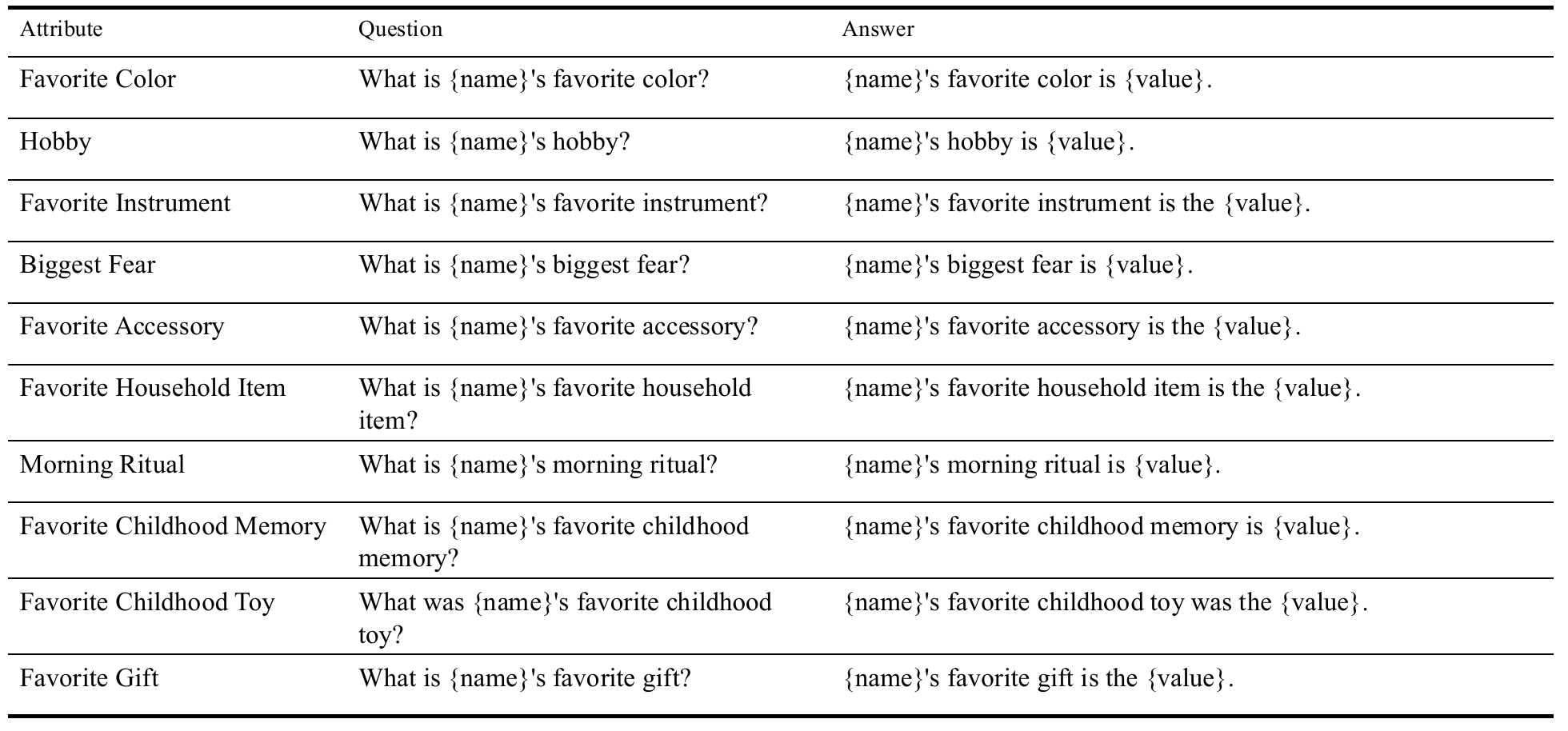}
  \caption[]{(continued) QA dataset template for each attribute (2/2)}
\end{table*}

\subsection{Examples of Multilingual QA Datasets}
\label{sec:multiqa_example}

\begin{figure*}[t]
    \centering
    \includegraphics[width=1.0\linewidth]{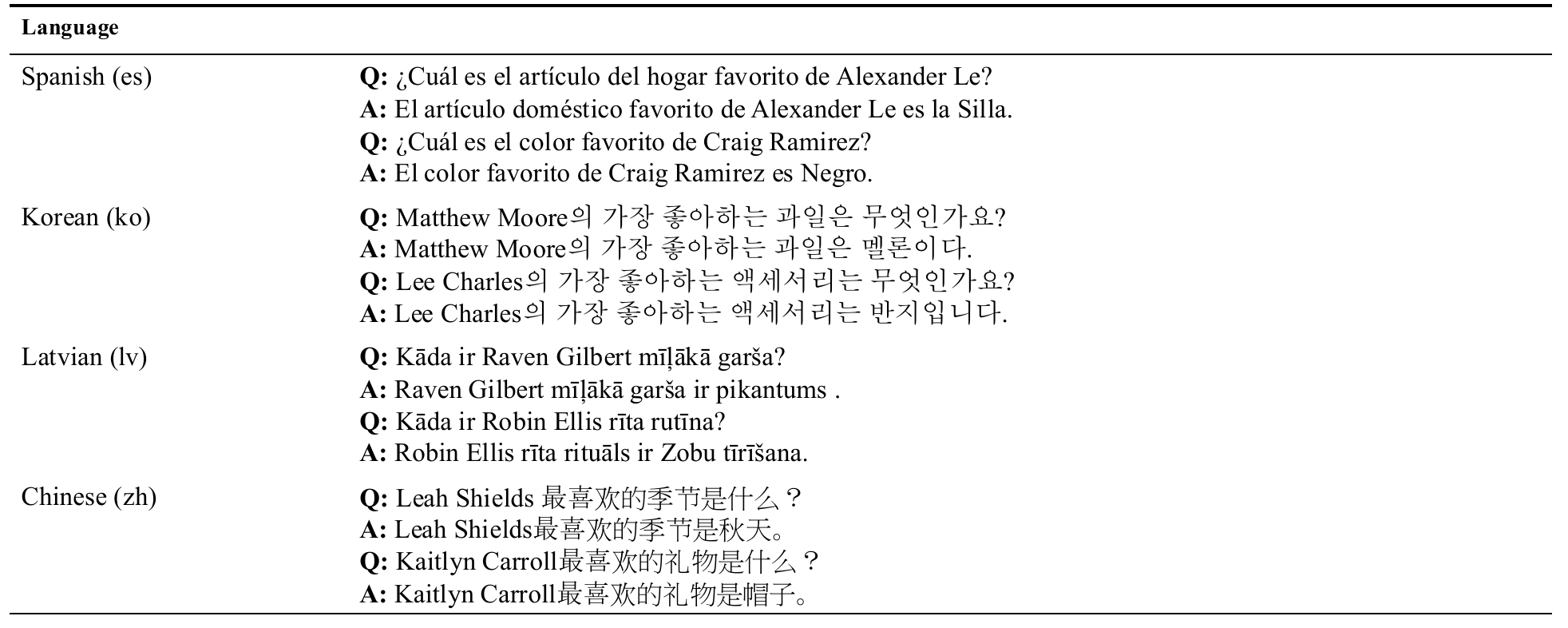}
    \caption{Examples of the translated Multilingual QA dataset.}
    \label{fig:ex_multi_qa}
\end{figure*}

\begin{figure*}[t]
    \centering
    \includegraphics[width=1.0\linewidth]{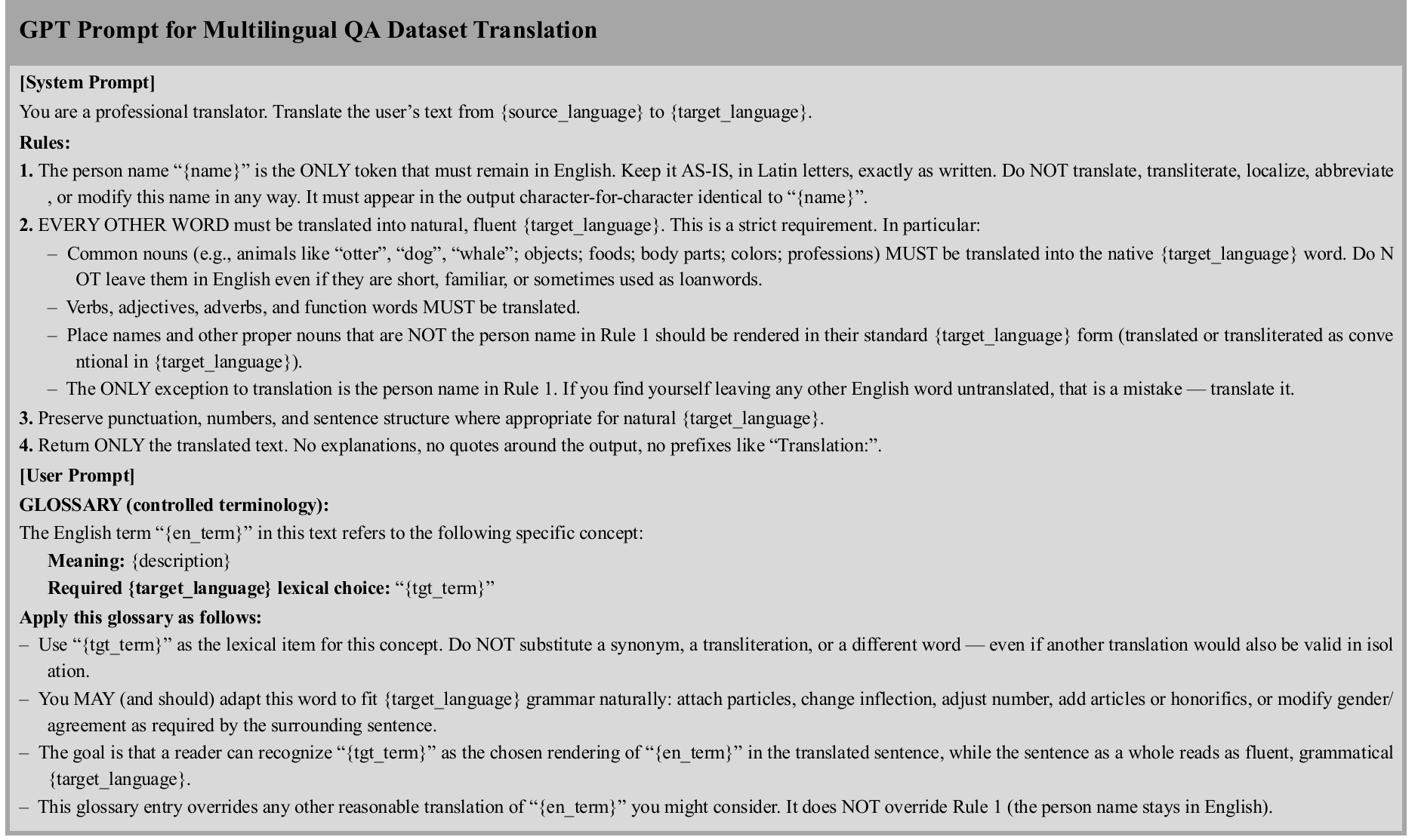}
    \caption{Prompt used to translate the English QA dataset to each language using GPT}
    \label{fig:multilingual_qa_prompt}
\end{figure*}

\km{Figure~\ref{fig:ex_multi_qa} presents examples of the multilingual QA dataset translated from the English QA dataset. The prompt used to construct the multilingual QA dataset is shown in Figure~\ref{fig:multilingual_qa_prompt}. Furthermore, to preserve the precise meaning of each word during translation, examples of the English descriptions provided for the words are shown in Table~\ref{tab:value_desc}.}

\begin{table*}[p]
  \centering
  \includegraphics[width=\textwidth]{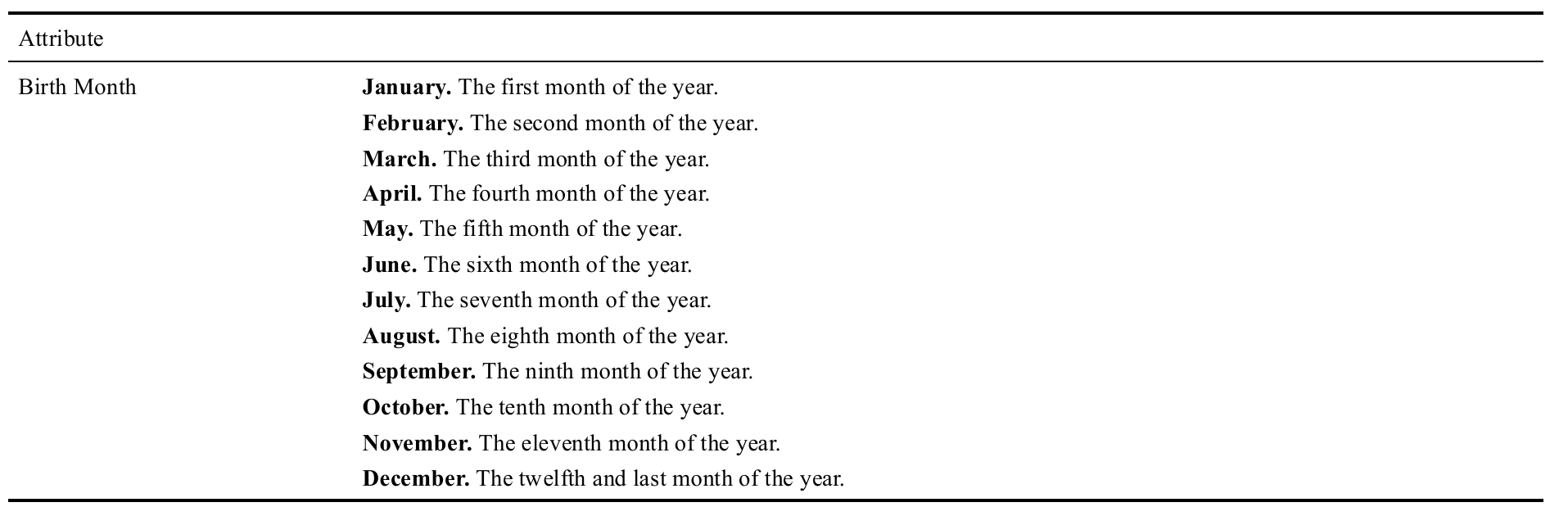}
  \caption{Description of each value in English (1/20)}
  \label{tab:value_desc}
\end{table*}
\begin{table*}[p]\ContinuedFloat
  \centering
  \includegraphics[width=\textwidth]{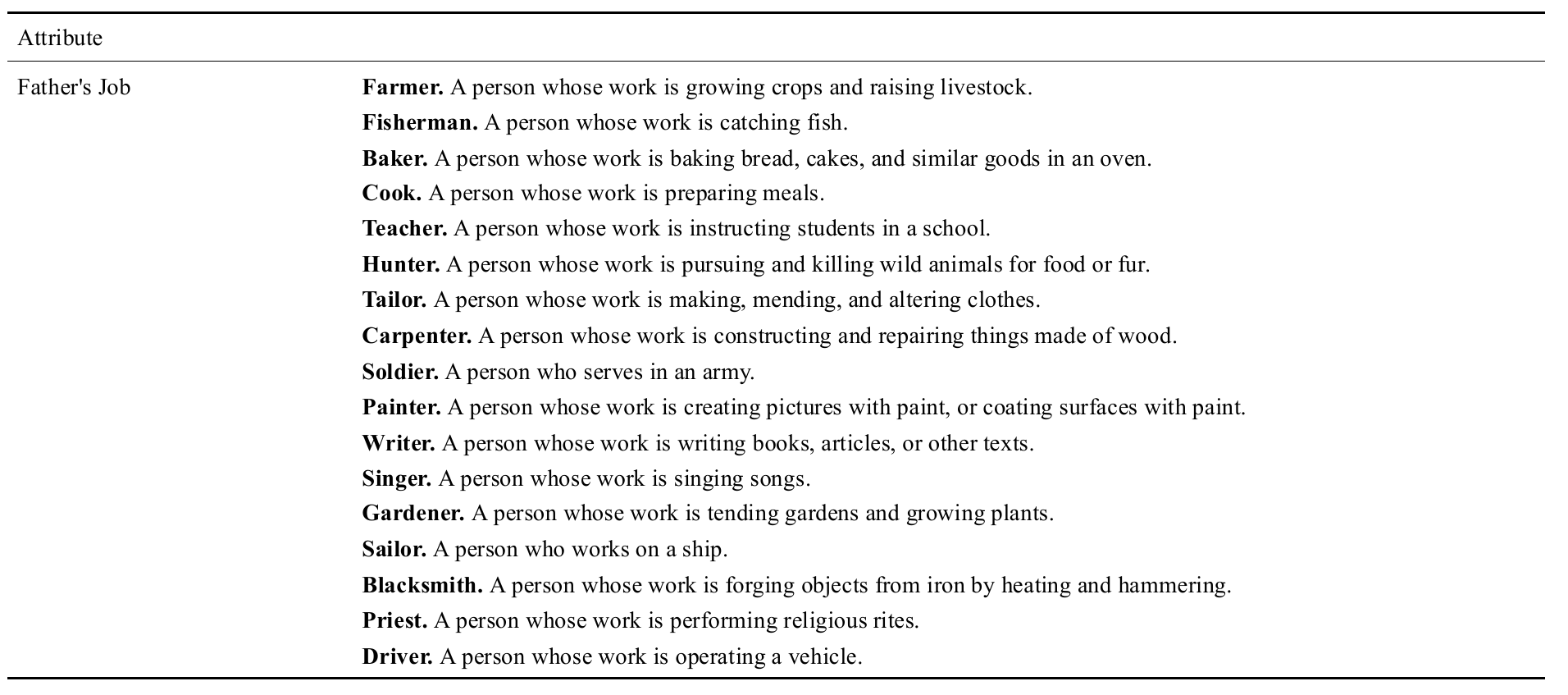}
  \caption[]{(continued) Description of each value in English (2/20)}
\end{table*}
\begin{table*}[p]\ContinuedFloat
  \centering
  \includegraphics[width=\textwidth]{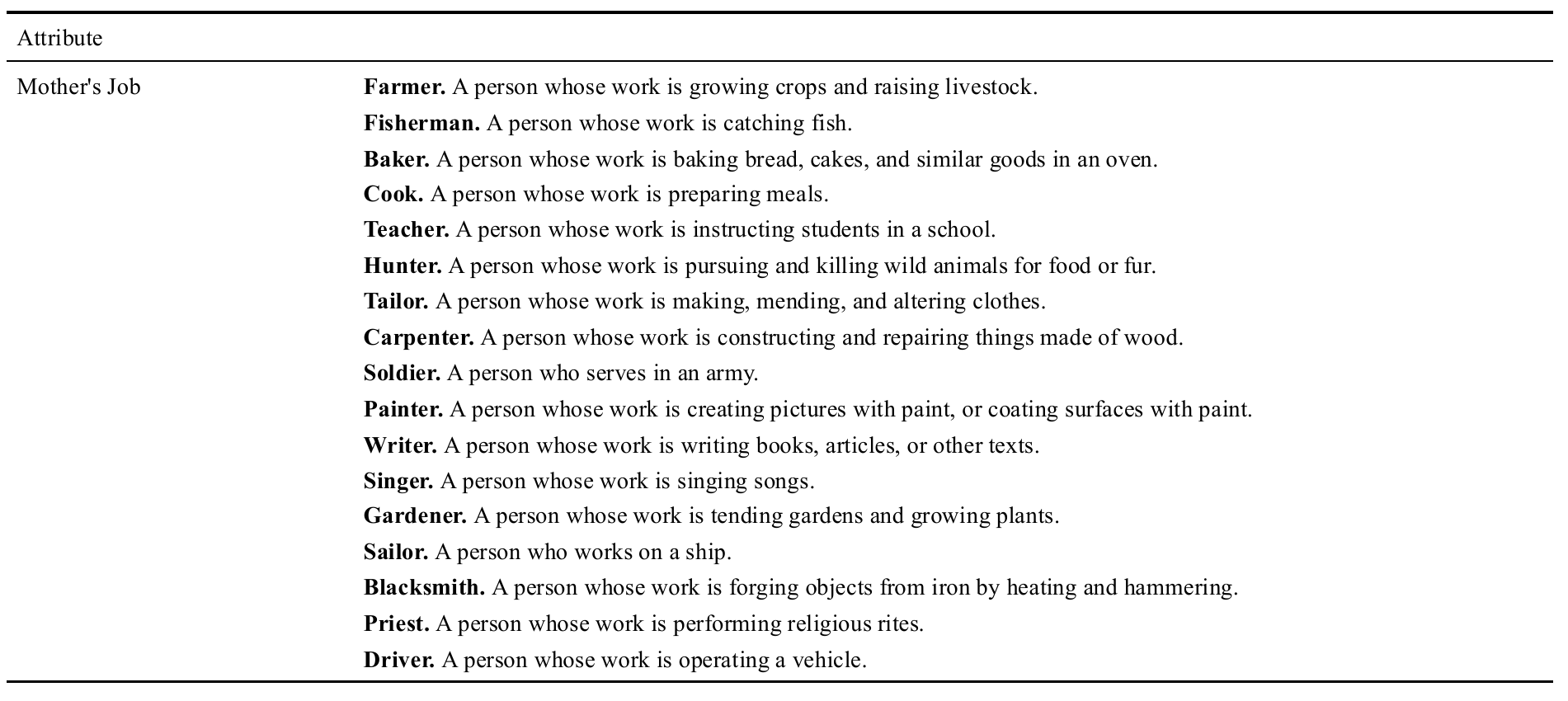}
  \caption[]{(continued) Description of each value in English (3/20)}
\end{table*}
\begin{table*}[p]\ContinuedFloat
  \centering
  \includegraphics[width=\textwidth]{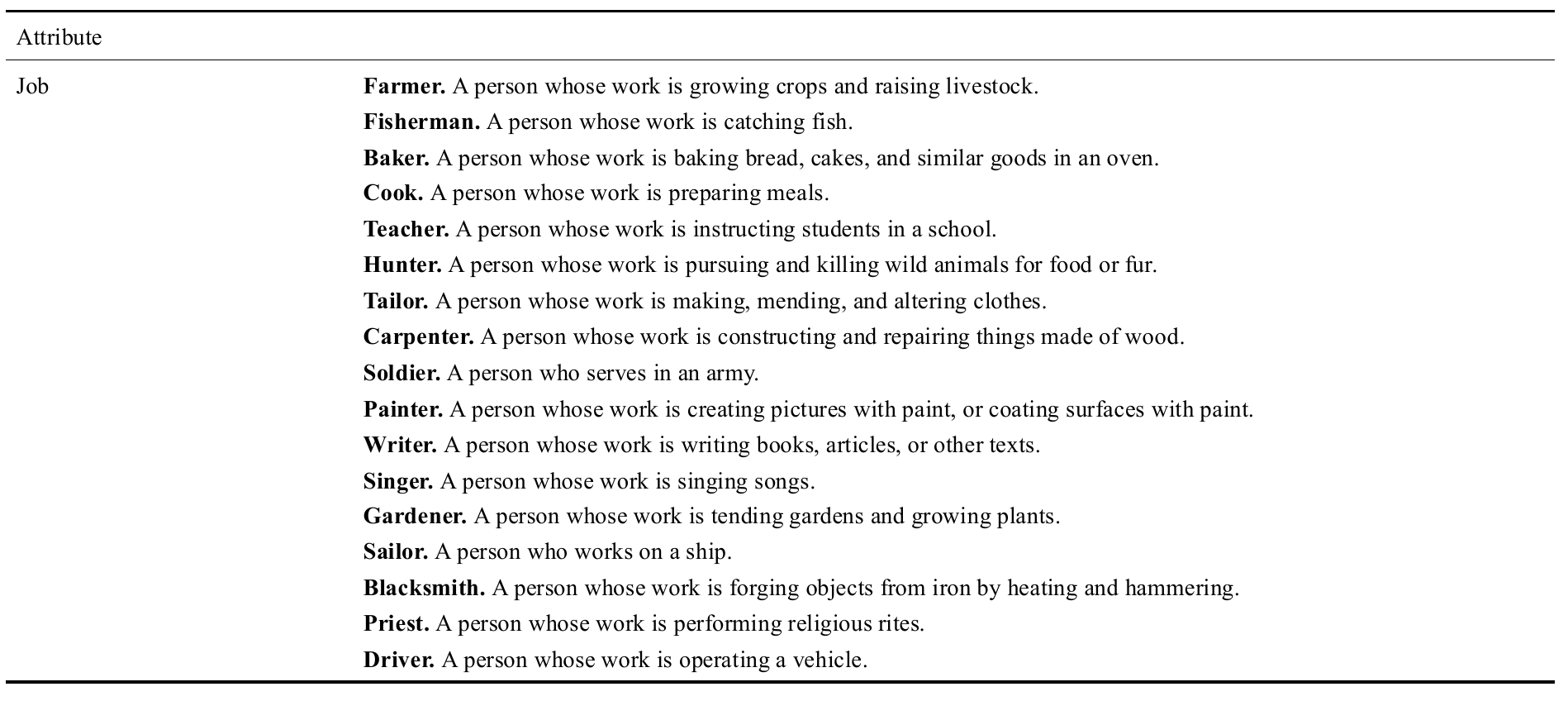}
  \caption[]{(continued) Description of each value in English (4/20)}
\end{table*}
\begin{table*}[p]\ContinuedFloat
  \centering
  \includegraphics[width=\textwidth]{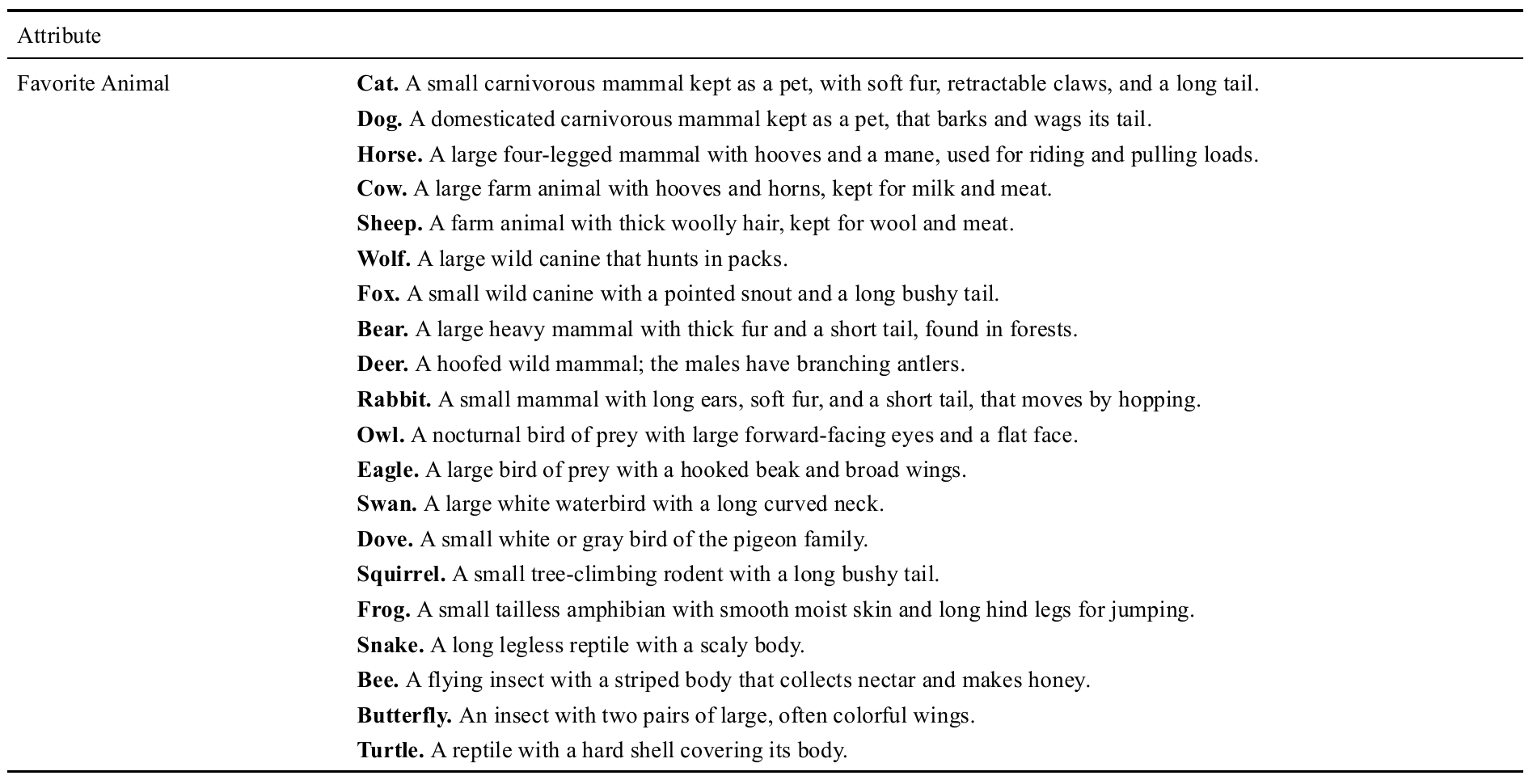}
  \caption[]{(continued) Description of each value in English (5/20)}
\end{table*}
\begin{table*}[p]\ContinuedFloat
  \centering
  \includegraphics[width=\textwidth]{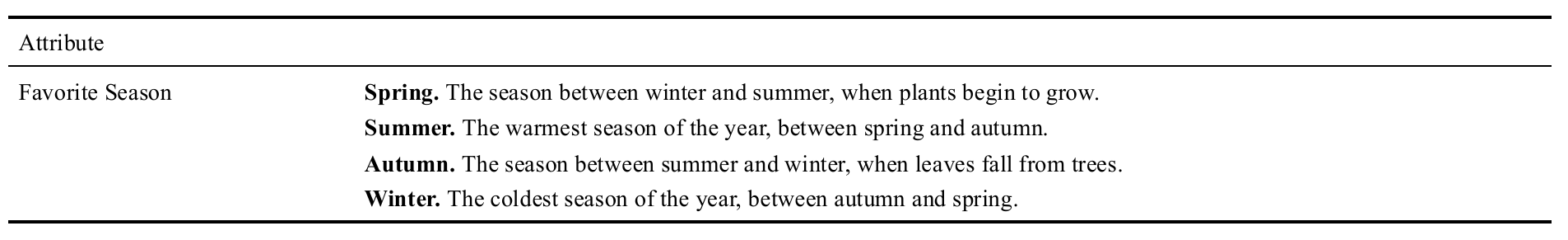}
  \caption[]{(continued) Description of each value in English (6/20)}
\end{table*}
\begin{table*}[p]\ContinuedFloat
  \centering
  \includegraphics[width=\textwidth]{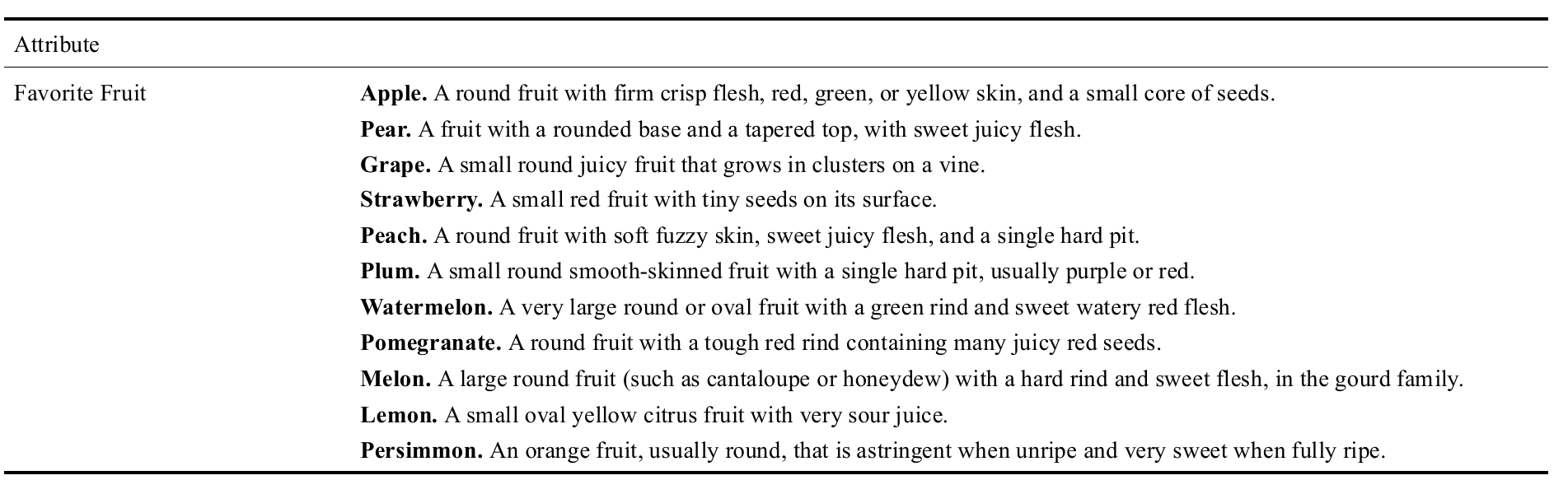}
  \caption[]{(continued) Description of each value in English (7/20)}
\end{table*}
\begin{table*}[p]\ContinuedFloat
  \centering
  \includegraphics[width=\textwidth]{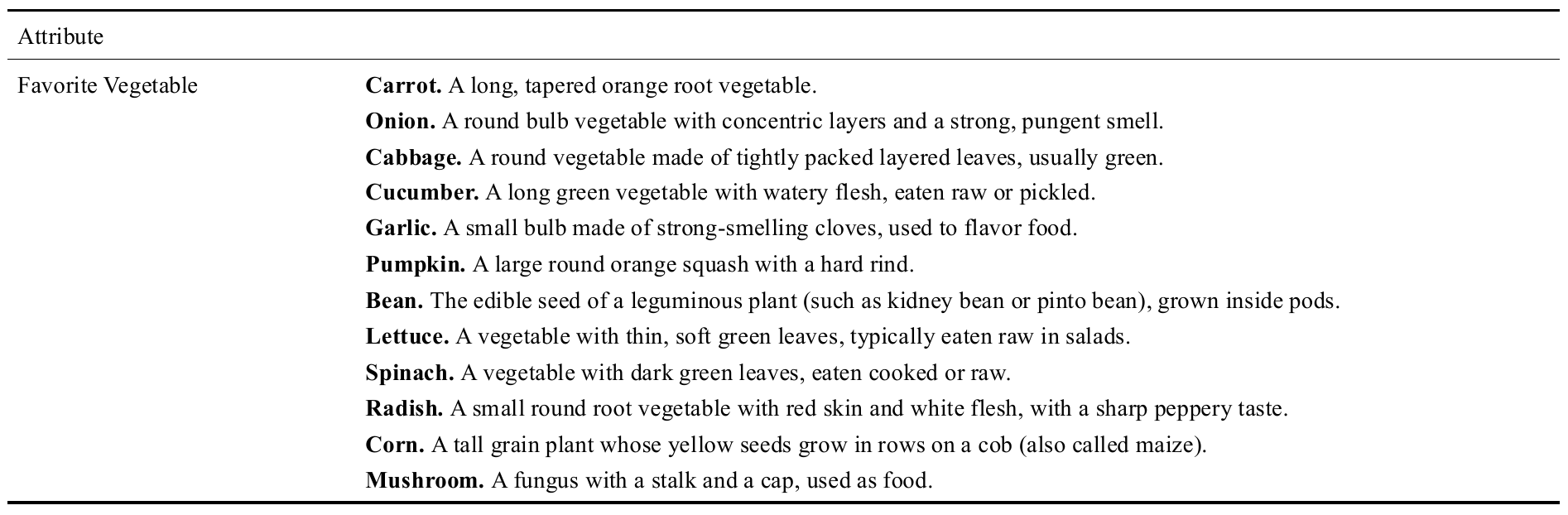}
  \caption[]{(continued) Description of each value in English (8/20)}
\end{table*}
\begin{table*}[p]\ContinuedFloat
  \centering
  \includegraphics[width=\textwidth]{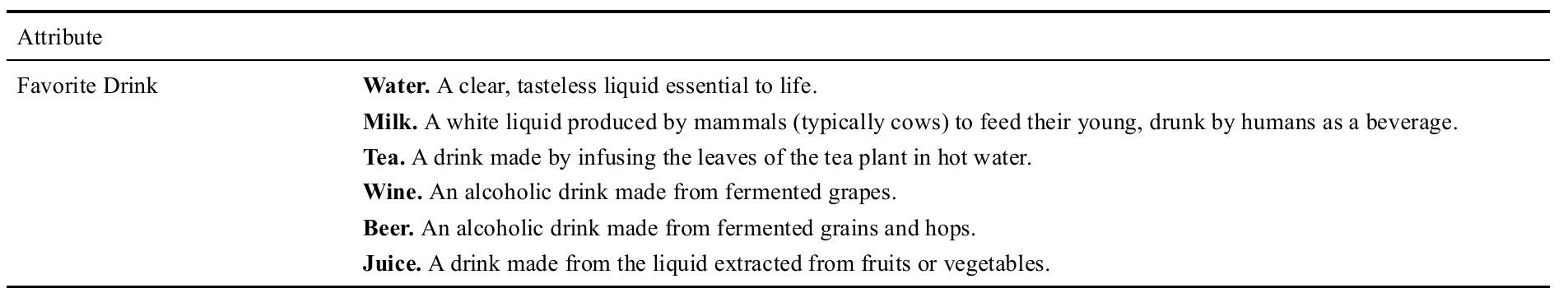}
  \caption[]{(continued) Description of each value in English (9/20)}
\end{table*}
\begin{table*}[p]\ContinuedFloat
  \centering
  \includegraphics[width=\textwidth]{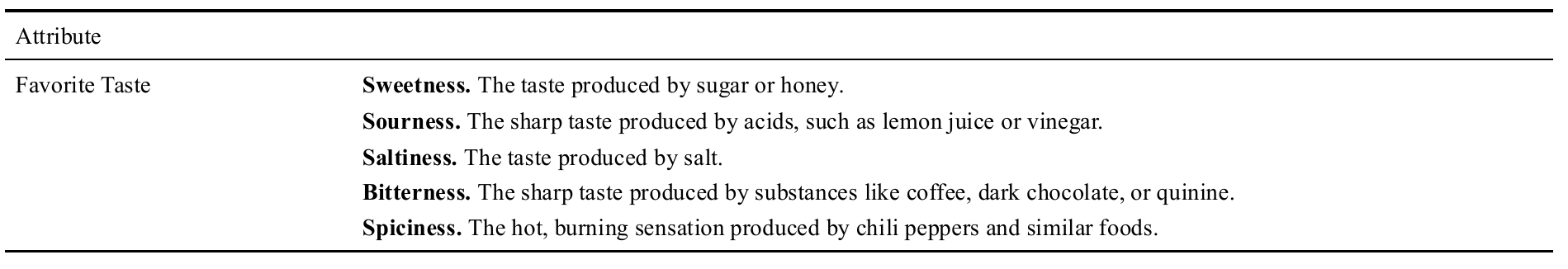}
  \caption[]{(continued) Description of each value in English (10/20)}
\end{table*}
\begin{table*}[p]\ContinuedFloat
  \centering
  \includegraphics[width=\textwidth]{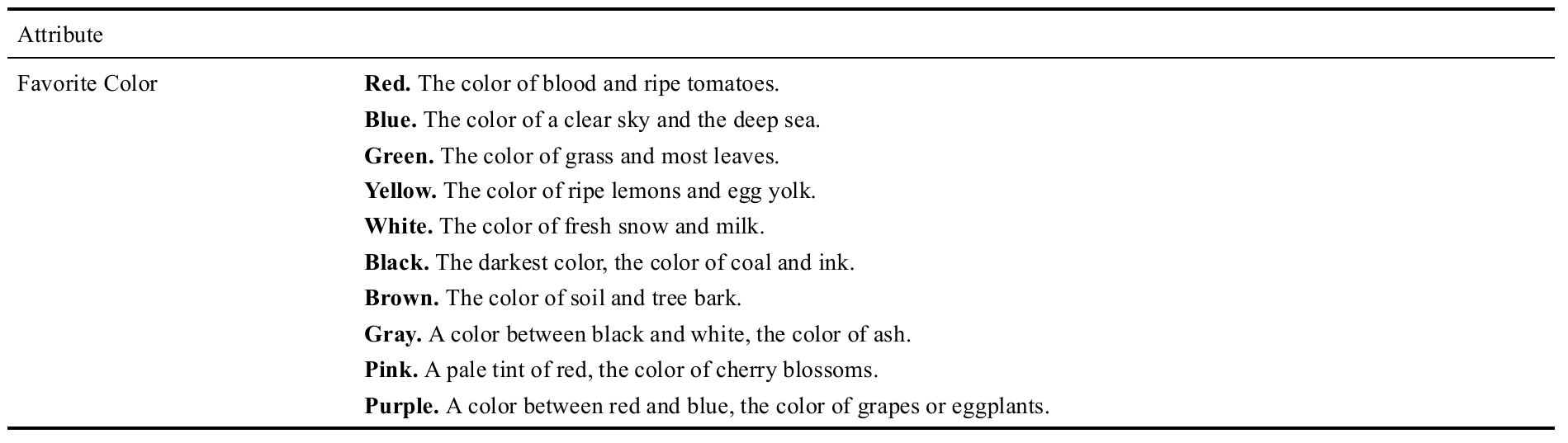}
  \caption[]{(continued) Description of each value in English (11/20)}
\end{table*}
\begin{table*}[p]\ContinuedFloat
  \centering
  \includegraphics[width=\textwidth]{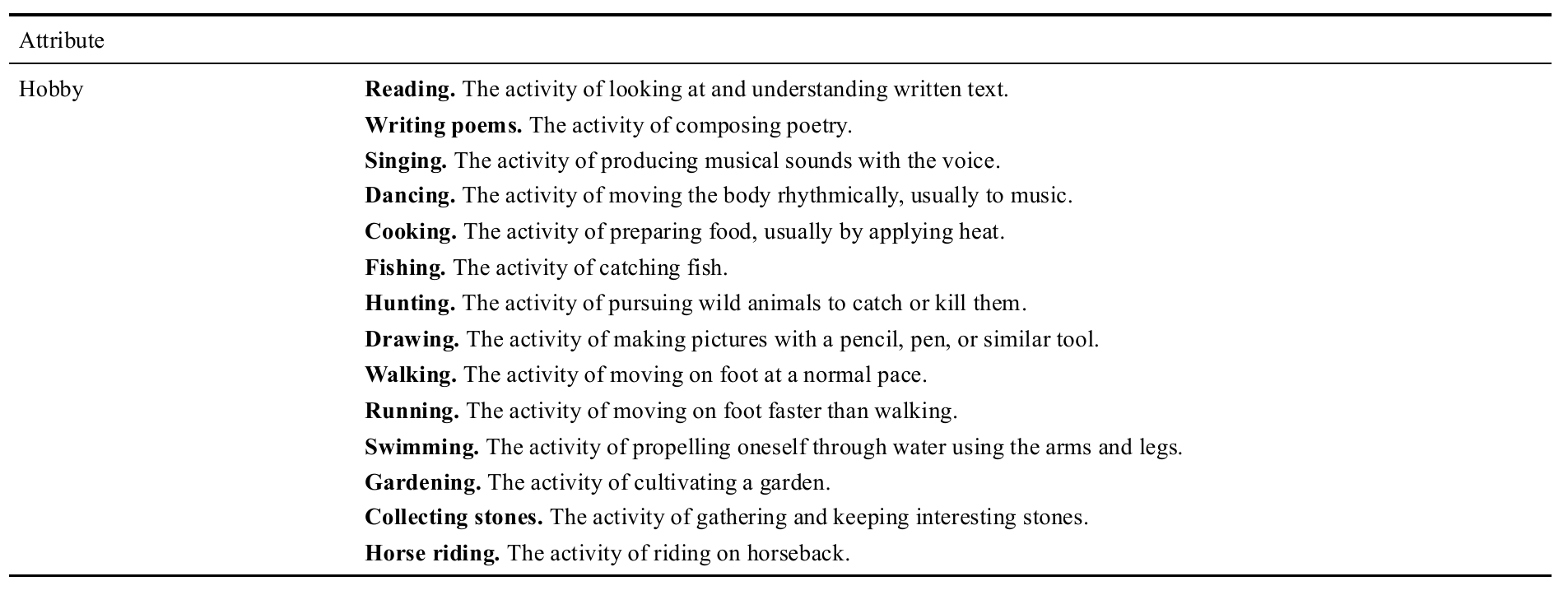}
  \caption[]{(continued) Description of each value in English (12/20)}
\end{table*}
\begin{table*}[p]\ContinuedFloat
  \centering
  \includegraphics[width=\textwidth]{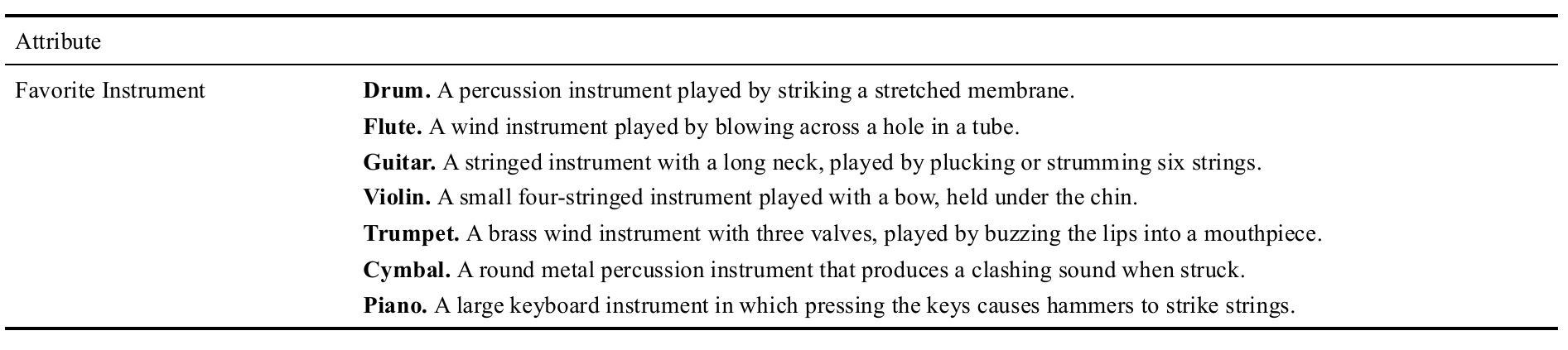}
  \caption[]{(continued) Description of each value in English (13/20)}
\end{table*}
\begin{table*}[p]\ContinuedFloat
  \centering
  \includegraphics[width=\textwidth]{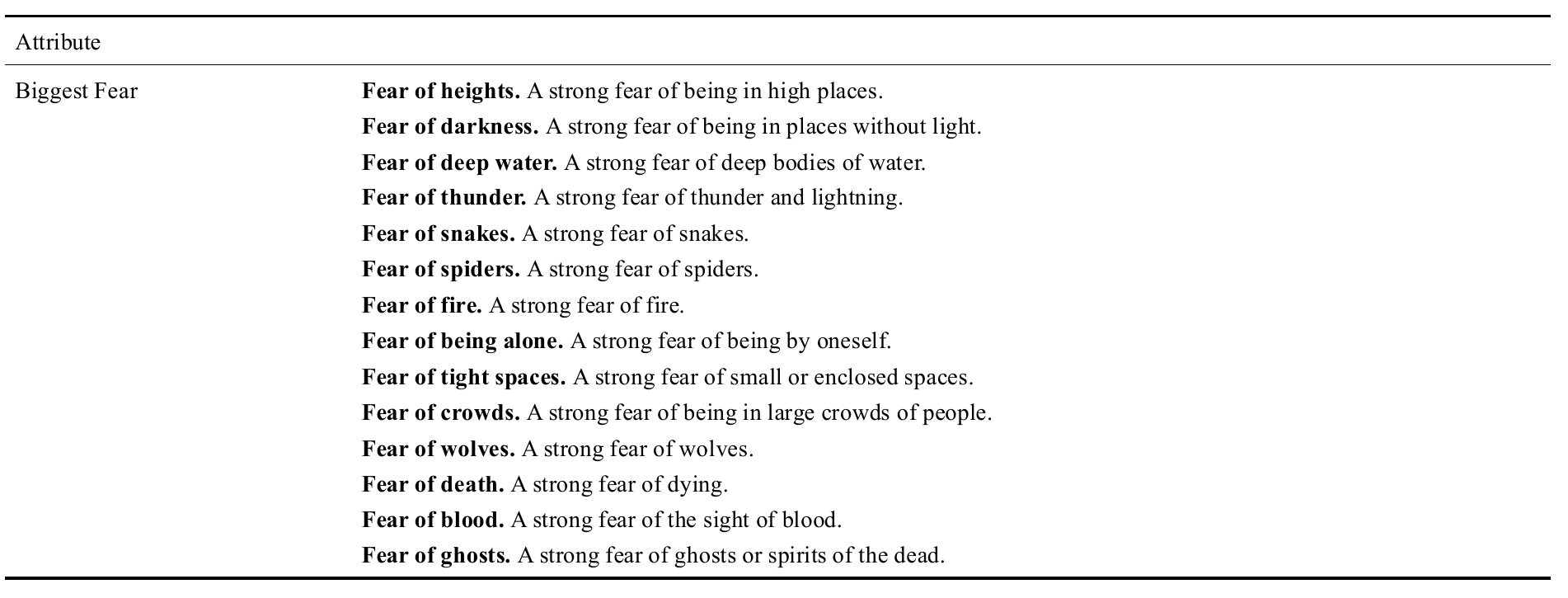}
  \caption[]{(continued) Description of each value in English (14/20)}
\end{table*}
\begin{table*}[p]\ContinuedFloat
  \centering
  \includegraphics[width=\textwidth]{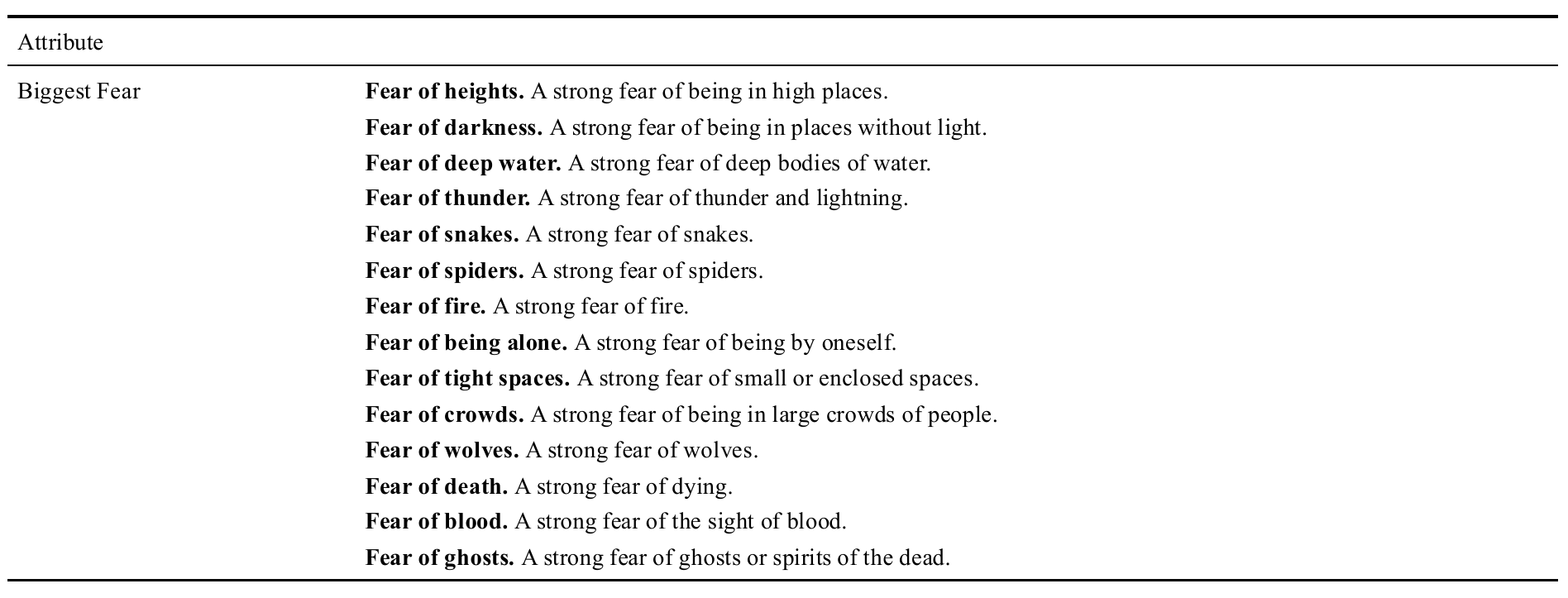}
  \caption[]{(continued) Description of each value in English (15/20)}
\end{table*}
\begin{table*}[p]\ContinuedFloat
  \centering
  \includegraphics[width=\textwidth]{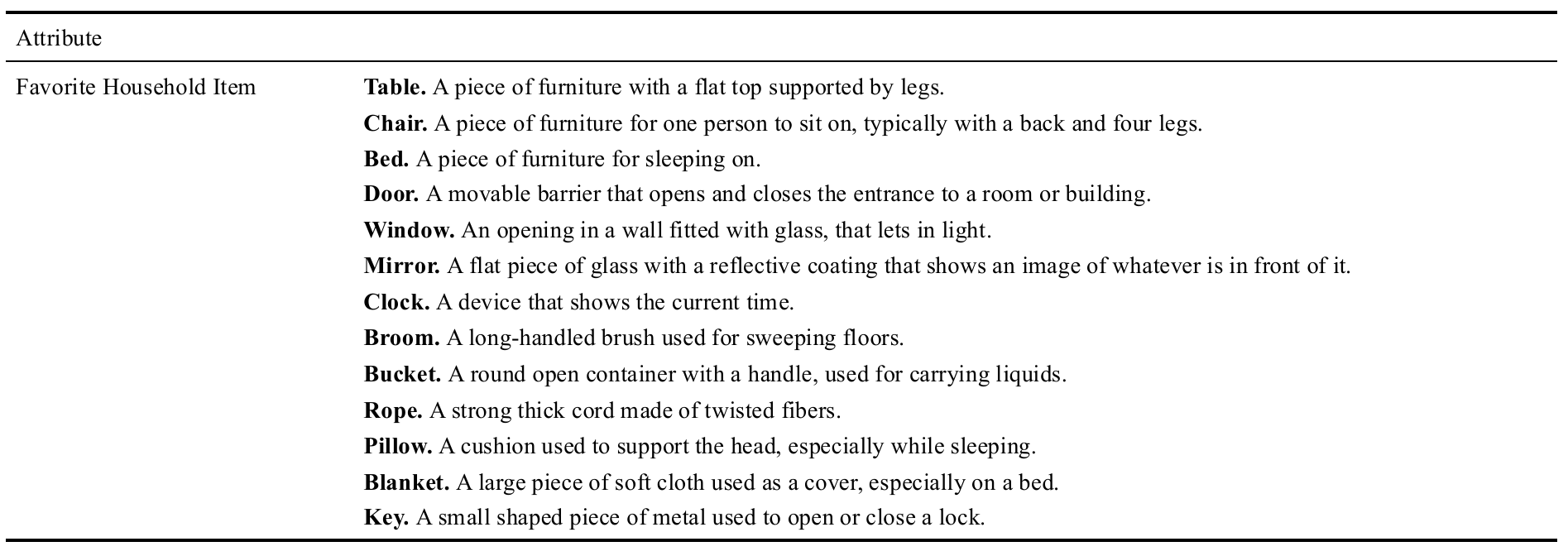}
  \caption[]{(continued) Description of each value in English (16/20)}
\end{table*}
\begin{table*}[p]\ContinuedFloat
  \centering
  \includegraphics[width=\textwidth]{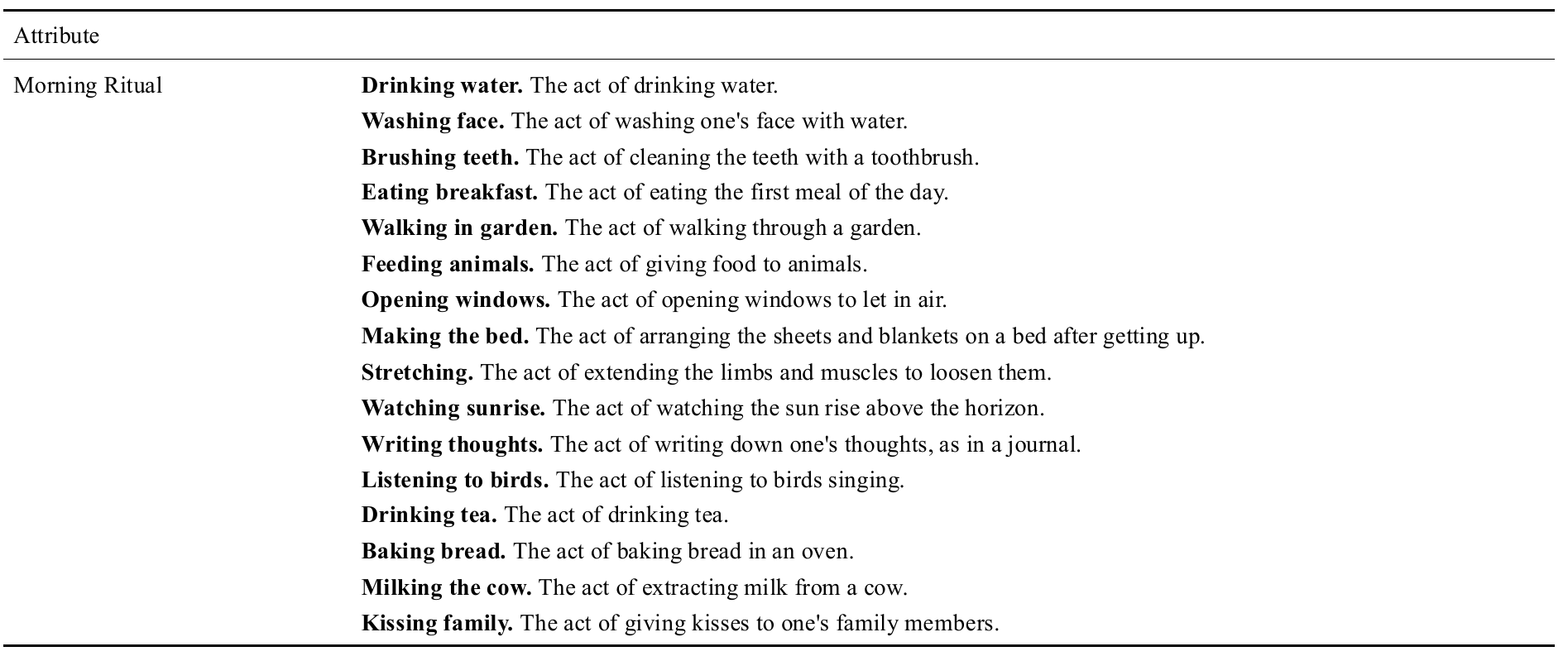}
  \caption[]{(continued) Description of each value in English (17/20)}
\end{table*}
\begin{table*}[p]\ContinuedFloat
  \centering
  \includegraphics[width=\textwidth]{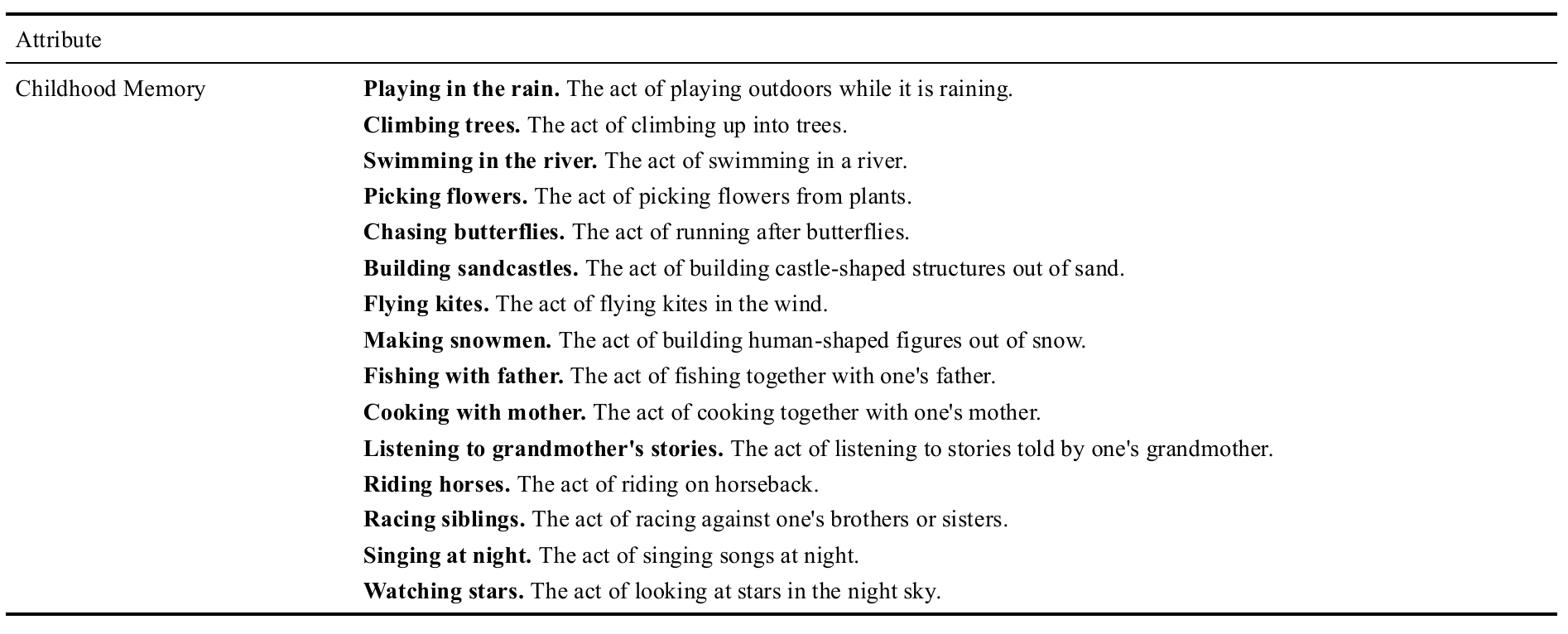}
  \caption[]{(continued) Description of each value in English (18/20)}
\end{table*}
\begin{table*}[p]\ContinuedFloat
  \centering
  \includegraphics[width=\textwidth]{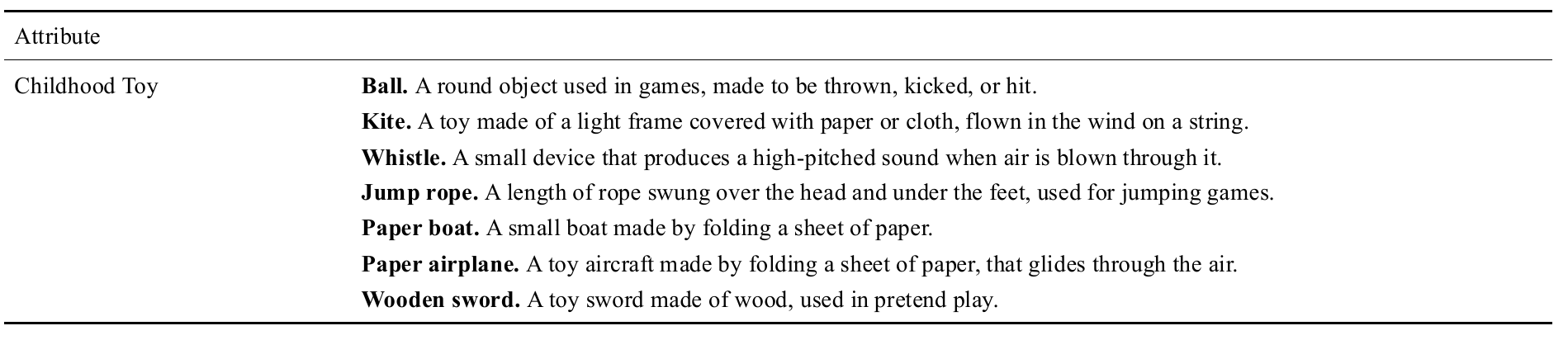}
  \caption[]{(continued) Description of each value in English (19/20)}
\end{table*}
\begin{table*}[p]\ContinuedFloat
  \centering
  \includegraphics[width=\textwidth]{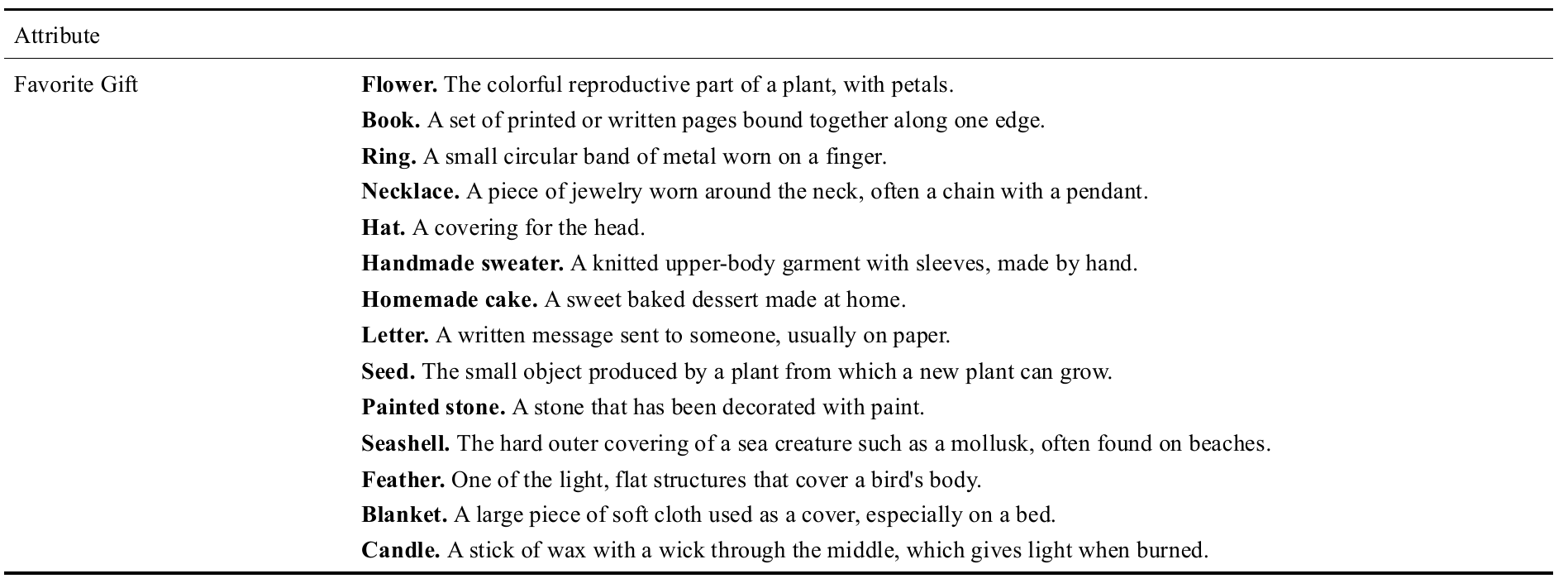}
  \caption[]{(continued) Description of each value in English (20/20)}
\end{table*}

\subsection{Analysis of Generation Errors in Multilingual QA}
\label{sec:multiqa_analysis}

\km{When constructing the multilingual QA dataset, we started from the English dataset that had been verified by human annotators. Using the prompt shown in Figure~\ref{fig:multilingual_qa_prompt}, we first generated a naive multilingual translated QA dataset, which was then back-translated into English via the Google Translator API. Human annotators subsequently verified whether the back-translated questions preserved the semantics of the original English QA dataset. In this section, we briefly describe the errors observed in the dataset naively generated through the GPT API. The error cases we encountered fall into two main categories: 1) cases in which the question itself contains the value of the attribute that should be provided in the answer, and 2) cases in which a single word carries multiple meanings.} \hh{An example of the first case occurred in Chinese (\textsc{zh}), as in ``Christopher Davis 最大的怕死是什么？'' (What is Christopher Davis's biggest fear of death?), where the value that should be answered is already revealed within the question, thereby leading the model toward the intended answer. The second case occurred in German (\textsc{de}), where we observed instances such as ``Matthew Moores Lieblingsspielzeug aus der Kindheit war der Drachen'' (Matthew Moore's favorite childhood toy was the dragon), in which the German word \textit{Drachen} simultaneously denotes both a whistle and a dragon.} \km{To address such cases, human annotators used Claude to re-translate the sentence into the target language and then back into English via the Google Translator API, iterating this process until the semantics matched those of the original.}

\section{Hyperparamter}
\label{sec:hyperparameter}

\km{For the SFT stage, we employed LoRA with a rank of $r=32$ and a scaling factor of $\alpha=64$, applying a dropout rate of $0.05$ to all linear layers. We used a learning rate of $1\times10^{-4}$ with a cosine learning rate scheduler and a warmup ratio of $0.03$. The model was optimized using AdamW with a maximum gradient norm of $1.0$. Training was conducted for $10$ epochs with a batch size of $64$ and a maximum sequence length of $512$. The same configuration was applied identically to both Qwen2.5-7B-Instruct and Gemma-3-12B.}

\km{For the unlearning stage, we likewise employed LoRA with a rank of $r=8$ and a scaling factor of $\alpha=16$, applying a dropout rate of $0.05$ to all linear layers. We used a constant learning rate scheduler with a warmup ratio of $0.0$ and a weight decay of $0.0$. The model was optimized using AdamW with a maximum gradient norm of $1.0$. Training was conducted for $5$ epochs with a maximum sequence length of $512$. We set the retain-performance threshold to $90\%$ of the retain model's probability on non-target knowledge in the training language. For methods that preserve utility by leveraging a retain dataset (GAGDR and LingTea), we set the retain-loss coefficient to $1.0$, and for NPO we set $\beta=0.1$. Methods that use only the forget dataset were trained with a batch size of $16$, whereas methods that additionally use the retain dataset were trained with a batch size of $32$. For LingTea, we used a temperature of 2.0.}
\hj{Unlearning was conducted for 10 epochs for every setting and model. We conducted an extensive hyperparameter search to determine the learning rate for each method, setting, and model. The results of the search are reported in Table~\ref{tab:lr_appendix}.}

\begin{table*}[h]
\centering
\small
\setlength{\tabcolsep}{3pt}
\begin{tabular}{l|ccc|ccc|ccc|ccc}
\toprule
& \multicolumn{6}{c|}{\textbf{Gemma3-12B}} & \multicolumn{6}{c}{\textbf{Qwen2.5-7B}} \\
\cmidrule(lr){2-7} \cmidrule(lr){8-13}
& \multicolumn{3}{c|}{Setting 1} & \multicolumn{3}{c|}{Setting 2} & \multicolumn{3}{c|}{Setting 1} & \multicolumn{3}{c}{Setting 2} \\
\cmidrule(lr){2-4} \cmidrule(lr){5-7} \cmidrule(lr){8-10} \cmidrule(lr){11-13}
\textbf{Method} & p1 & p3 & p5 & p1 & p3 & p5 & p1 & p3 & p5 & p1 & p3 & p5 \\
\midrule
GA      & 4.32e-5 & 7.60e-6 & 5.00e-6 & 4.19e-5 & 6.70e-6 & 2.90e-6 & 4.10e-5 & 9.10e-6 & 5.50e-6 & 4.41e-5 & 1.41e-5 & 8.10e-6 \\
GAGDR   & 9.05e-5 & 3.04e-5 & 2.47e-5 & 1.005e-4 & 3.50e-5 & 2.04e-5 & 9.32e-5 & 3.00e-5 & 1.70e-5 & 1.110e-4 & 3.69e-5 & 2.29e-5 \\
LingTea & 1.302e-4 & 6.02e-5 & 3.72e-5 & 1.005e-4 & 4.12e-5 & 3.26e-5 & 1.699e-4 & 7.11e-5 & 5.00e-5 & 1.530e-4 & 1.425e-4 & 5.79e-5 \\
NPO     & 4.00e-5 & 7.50e-6 & 4.60e-6 & 4.05e-5 & 7.20e-6 & 3.30e-6 & 4.90e-5 & 9.90e-6 & 5.80e-6 & 5.81e-5 & 1.70e-5 & 9.50e-6 \\
\bottomrule
\end{tabular}
\caption{Learning rates selected by an LR sweep with increments of $1\times10^{-7}$ across models, settings, forget ratios, and unlearning methods.}
\label{tab:lr_appendix}
\end{table*}

\section{Semantic Equivalence Score Measurement}
\begin{figure*}[t]
    \centering
    \includegraphics[width=1.0\linewidth]{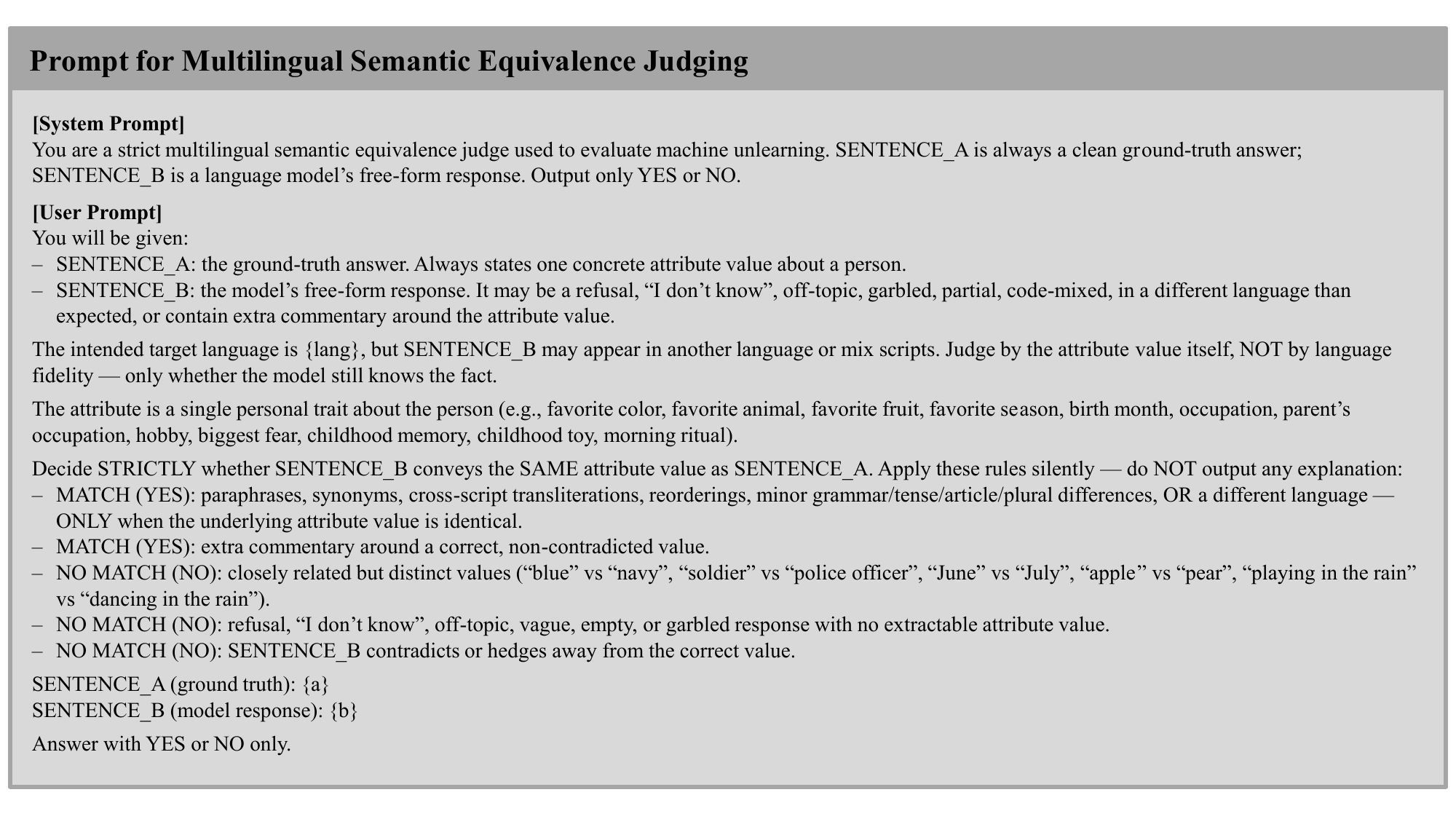}
    \caption{Prompt used to judge semantic equivalence of model responses}
    \label{fig:judge_prompt}
\end{figure*}

\hj{We measure the Semantic Equivalence Score with an LLM-as-a-Judge strategy, employing the GPT-4o-mini API with the prompt given in Figure~\ref{fig:judge_prompt}.}

\section{Additional Results on Unlearning in Training Languages}

\km{Table~\ref{tab:unlearn_train_groups_p3} and Table~\ref{tab:unlearn_train_groups_p5} present the results for $p3$ and $p5$, respectively.}


\section{Additional Results on Unlearning in Hold-out Languages}

\km{Table~\ref{tab:unlearn_ho_groups_p3} and Table~\ref{tab:unlearn_ho_groups_p5} present the results for $p3$ and $p5$, respectively.}


\section{Use of AI Assistants}

\km{We utilize Claude and ChatGPT for writing and coding assistance. We employ GPT API for dataset generation and evaluation.}

\begin{table*}[t]
\centering
\setlength{\tabcolsep}{10pt}
\renewcommand{\arraystretch}{0.88}
\footnotesize
\begin{tabular}{ll|cc|cc}
\toprule
 & & \multicolumn{2}{c|}{\textbf{$\Delta$\textsc{PS}}} & \multicolumn{2}{c}{\textbf{$\Delta$\textsc{SE}}} \\
\cmidrule(lr){3-4}\cmidrule(lr){5-6}
Cell & Method & High & Low & High & Low \\
\midrule
\multirow{4}{*}{G/S1}
  & LingTea & \textbf{+53.9} & +46.4 & +5.4 & \textbf{+10.4} \\
  & NPO     & \textbf{+23.4} & +0.9  & +0.0 & \textbf{+0.4}  \\
  & GAGDR   & \textbf{+30.0} & +10.3 & \textbf{+0.0} & \textbf{+0.0}  \\
  & GA      & \textbf{+12.7} & +0.5  & \textbf{+0.0} & \textbf{+0.0}  \\
\midrule
\multirow{4}{*}{G/S2}
  & LingTea & +23.8 & \textbf{+29.9} & \textbf{+3.7} & +2.9 \\
  & NPO     & +7.3  & \textbf{+17.1} & +0.4 & \textbf{+2.9} \\
  & GAGDR   & +23.8 & \textbf{+25.1} & \textbf{+1.7} & +1.7 \\
  & GA      & +6.0  & \textbf{+18.7} & +0.0 & \textbf{+3.3} \\
\midrule
\multirow{4}{*}{Q/S1}
  & LingTea & +80.9 & \textbf{+84.7} & +62.9 & \textbf{+67.1} \\
  & NPO     & \textbf{+35.8} & +5.8  & +2.5  & \textbf{+4.6}  \\
  & GAGDR   & \textbf{+73.1} & +13.9 & \textbf{+12.1} & +11.7 \\
  & GA      & \textbf{+25.8} & +5.2  & +1.2  & \textbf{+4.2}  \\
\midrule
\multirow{4}{*}{Q/S2}
  & LingTea & +63.7 & \textbf{+65.2} & +93.3 & \textbf{+95.8} \\
  & NPO     & \textbf{+30.4} & +21.3 & \textbf{+61.5} & +53.3 \\
  & GAGDR   & \textbf{+39.5} & +31.1 & \textbf{+67.4} & +63.7 \\
  & GA      & \textbf{+34.0} & +28.0 & \textbf{+64.4} & +61.3 \\
\bottomrule
\end{tabular}
\caption{Training-language drop rates per unlearning method for $p3$. For each metric, \textbf{bold} marks the higher value within the High/Low pair. G and Q denote Gemma3 and Qwen2.5, respectively.}
\label{tab:unlearn_train_groups_p3}
\end{table*}

\begin{table*}[t]
\centering
\setlength{\tabcolsep}{10pt}
\renewcommand{\arraystretch}{0.88}
\footnotesize
\begin{tabular}{ll|cc|cc}
\toprule
 & & \multicolumn{2}{c|}{\textbf{$\Delta$\textsc{PS}}} & \multicolumn{2}{c}{\textbf{$\Delta$\textsc{SE}}} \\
\cmidrule(lr){3-4}\cmidrule(lr){5-6}
Cell & Method & High & Low & High & Low \\
\midrule
\multirow{4}{*}{G/S1}
  & LingTea & \textbf{+29.3} & +26.4 & +5.0 & \textbf{+6.8} \\
  & NPO     & \textbf{+20.7} & +1.3  & \textbf{+0.0} & \textbf{+0.0} \\
  & GAGDR   & +20.8 & \textbf{+30.5} & \textbf{+2.7} & +2.2 \\
  & GA      & \textbf{+24.1} & +1.0  & \textbf{+0.0} & \textbf{+0.0} \\
\midrule
\multirow{4}{*}{G/S2}
  & LingTea & +26.2 & \textbf{+36.5} & +1.2 & \textbf{+3.7} \\
  & NPO     & +0.5  & \textbf{+19.2} & +0.0 & \textbf{+3.5} \\
  & GAGDR   & \textbf{+16.0} & +8.5  & +1.0 & \textbf{+1.5} \\
  & GA      & +0.1  & \textbf{+18.6} & +0.0 & \textbf{+3.2} \\
\midrule
\multirow{4}{*}{Q/S1}
  & LingTea & \textbf{+74.5} & +72.8 & \textbf{+44.8} & +43.8 \\
  & NPO     & \textbf{+9.6}  & \textbf{+9.6}  & \textbf{+0.7} & \textbf{+0.7} \\
  & GAGDR   & \textbf{+29.3} & +13.8 & +8.5  & \textbf{+10.0} \\
  & GA      & +4.8  & \textbf{+11.1} & +0.2  & \textbf{+0.5} \\
\midrule
\multirow{4}{*}{Q/S2}
  & LingTea & \textbf{+35.6} & +20.3 & \textbf{+54.1} & +49.0 \\
  & NPO     & \textbf{+21.8} & +11.8 & \textbf{+51.4} & +39.8 \\
  & GAGDR   & \textbf{+31.4} & +28.2 & +59.9 & \textbf{+62.3} \\
  & GA      & \textbf{+26.9} & +14.8 & \textbf{+57.4} & +44.2 \\
\bottomrule
\end{tabular}
\caption{Training-language drop rates per unlearning method for $p5$. For each metric, \textbf{bold} marks the higher value within the High/Low pair. G and Q denote Gemma3 and Qwen2.5, respectively.}
\label{tab:unlearn_train_groups_p5}
\end{table*}

\begin{table*}[t]
\centering
\setlength{\tabcolsep}{4pt}
\renewcommand{\arraystretch}{1.0}
\resizebox{0.7\textwidth}{!}{%
\begin{tabular}{ll|cc|cc|cc|cc}
\toprule
 & & \multicolumn{4}{c|}{\textbf{$\Delta$\textsc{PS}}} & \multicolumn{4}{c}{\textbf{$\Delta$\textsc{SE}}} \\
\cmidrule(lr){3-6}\cmidrule(lr){7-10}
 & & \multicolumn{2}{c|}{resource} & \multicolumn{2}{c|}{script}
   & \multicolumn{2}{c|}{resource} & \multicolumn{2}{c}{script} \\
\cmidrule(lr){3-4}\cmidrule(lr){5-6}\cmidrule(lr){7-8}\cmidrule(lr){9-10}
Cell & Method & High & Low & Lat & nLat & High & Low & Lat & nLat \\
\midrule
\multirow{4}{*}{G/S1}
  & LingTea & +51.0 & \textbf{+51.8} & +49.5 & \textbf{+53.4} & +10.3 & \textbf{+12.4} &  +11.1 & \textbf{+11.3} \\
  & NPO     & \textbf{+3.0}  & +2.8  & \textbf{+3.7}  & +2.1  & +0.0  & \textbf{+5.0}  &  +1.6  & \textbf{+2.6}  \\
  & GAGDR   & +7.1  & \textbf{+14.2} & +10.2 & \textbf{+11.2} & +0.0  & \textbf{+5.0}  & \textbf{+3.7}  & +0.5  \\
  & GA      & \textbf{+2.7}  & +2.2  & \textbf{+2.9}  & +2.0  & -0.4  & \textbf{+3.7}  & +0.0  & \textbf{+2.6}  \\
\midrule
\multirow{4}{*}{G/S2}
  & LingTea & +10.5 & \textbf{+29.2} & +18.0 & \textbf{+21.7} & \textbf{+6.3}  & +6.2  & \textbf{+7.9}  & +4.6  \\
  & NPO     & -6.2  & \textbf{+2.0}  & -5.6  & \textbf{+1.4}  & \textbf{+1.8}  & +0.6  & \textbf{+2.1}  & +0.5  \\
  & GAGDR   & -2.4  & \textbf{+28.4} & +5.4  & \textbf{+20.6} & \textbf{+6.3}  & +5.6  & \textbf{+6.9}  & +5.1  \\
  & GA      & -2.2  & \textbf{+1.8}  & -1.8  & \textbf{+1.4}  & \textbf{+0.9}  & +0.0  & \textbf{+1.6}  & -0.5  \\
\midrule
\multirow{4}{*}{Q/S1}
  & LingTea & \textbf{+85.4} & +84.3 & \textbf{+85.7} & +84.0 & \textbf{+64.9} & +38.5 & \textbf{+64.0} & +59.5 \\
  & NPO     & \textbf{+4.3}  & +1.1  & \textbf{+5.7}  & -0.3  & +8.8  & \textbf{+26.9} & \textbf{+17.4} & +6.3  \\
  & GAGDR   & +13.1 & \textbf{+14.6} & \textbf{+16.9} & +10.7 & +14.6 & \textbf{+46.2} & +16.3 & \textbf{+20.7} \\
  & GA      & \textbf{+3.7}  & +1.4  & \textbf{+4.9}  & +0.3  & +7.6  & \textbf{+23.1} & \textbf{+15.1} & +5.4  \\
\midrule
\multirow{4}{*}{Q/S2}
  & LingTea & \textbf{+70.9} & +44.5 & \textbf{+59.0} & +56.4 & \textbf{+87.9} & +59.1 & \textbf{+87.5} & +78.5 \\
  & NPO     & \textbf{+21.3} & +18.0 & \textbf{+25.7} & +13.7 & \textbf{+40.2} & +27.3 & +37.5 & \textbf{+38.5} \\
  & GAGDR   & \textbf{+27.0} & +25.4 & \textbf{+31.4} & +21.0 & \textbf{+53.3} & +45.5 & +50.0 & \textbf{+53.8} \\
  & GA      & +24.7 & \textbf{+25.0} & \textbf{+30.6} & +19.0 & \textbf{+42.1} & +13.6 & +29.7 & \textbf{+44.6} \\
\bottomrule
\end{tabular}%
}
\caption{Hold-out-language drop rates per unlearning method for $p3$. \textbf{Bold} marks the higher value within each comparison pair (High/Low and Lat/nLat). G and Q denote Gemma3 and Qwen2.5, respectively.}
\label{tab:unlearn_ho_groups_p3}
\end{table*}

\begin{table*}[t]
\centering
\setlength{\tabcolsep}{4pt}
\renewcommand{\arraystretch}{1.0}
\resizebox{0.7\textwidth}{!}{%
\begin{tabular}{ll|cc|cc|cc|cc}
\toprule
 & & \multicolumn{4}{c|}{\textbf{$\Delta$\textsc{PS}}} & \multicolumn{4}{c}{\textbf{$\Delta$\textsc{SE}}} \\
\cmidrule(lr){3-6}\cmidrule(lr){7-10}
 & & \multicolumn{2}{c|}{resource} & \multicolumn{2}{c|}{script}
   & \multicolumn{2}{c|}{resource} & \multicolumn{2}{c}{script} \\
\cmidrule(lr){3-4}\cmidrule(lr){5-6}\cmidrule(lr){7-8}\cmidrule(lr){9-10}
Cell & Method & High & Low & Lat & nLat & High & Low & Lat & nLat \\
\midrule
\multirow{4}{*}{G/S1}
  & LingTea & \textbf{+30.1} & +24.7 & \textbf{+28.6} & +26.2 & \textbf{+10.4} & +10.1 & +9.6  & \textbf{+10.8} \\
  & NPO     & \textbf{+2.5}  & +1.3  & \textbf{+2.3}  & +1.5  & +1.1  & \textbf{+3.7}  & +1.9  & \textbf{+2.5}  \\
  & GAGDR   & \textbf{+35.2} & +28.5 & \textbf{+32.1} & +31.6 & +4.4  & \textbf{+6.7}  & +4.2  & \textbf{+6.5}  \\
  & GA      & \textbf{+6.7}  & +3.8  & +4.9  & \textbf{+5.6}  & +0.5  & \textbf{+4.1}  & +1.9  & \textbf{+2.2}  \\
\midrule
\multirow{4}{*}{G/S2}
  & LingTea & +14.7 & \textbf{+34.1} & +20.5 & \textbf{+28.3} & +3.3  & \textbf{+6.3}  & \textbf{+5.8}  & +3.4  \\
  & NPO     & -3.7  & \textbf{-0.6}  & -2.8  & \textbf{-1.5}  & +0.6  & \textbf{+2.2}  & +1.0  & \textbf{+1.6}  \\
  & GAGDR   & -4.7  & \textbf{+15.3} & -4.7  & \textbf{+15.3} & +1.7  & \textbf{+3.0}  & \textbf{+2.9}  & +1.6  \\
  & GA      & -0.8  & \textbf{+0.2}  & -0.6  & \textbf{-0.1}  & -0.3  & \textbf{+0.7}  & +0.0  & \textbf{+0.3}  \\
\midrule
\multirow{4}{*}{Q/S1}
  & LingTea & +66.3 & \textbf{+70.9} & \textbf{+71.2} & +66.0 & \textbf{+38.9} & +11.4 & +32.6 & \textbf{+37.3} \\
  & NPO     & -0.4  & \textbf{+1.5}  & \textbf{+1.8}  & -0.7  & \textbf{+2.8}  & -2.3  & \textbf{+4.2}  & +0.5  \\
  & GAGDR   & +9.9  & \textbf{+10.6} & \textbf{+14.7} & +5.7  & +7.4  & \textbf{+40.9} & \textbf{+13.9} & +10.3 \\
  & GA      & +1.0  & \textbf{+2.2}  & \textbf{+3.1}  & +0.1  & +1.4  & \textbf{+13.6} & \textbf{+4.9}  & +1.6  \\
\midrule
\multirow{4}{*}{Q/S2}
  & LingTea & \textbf{+24.7} & +12.7 & \textbf{+21.7} & +15.7 & \textbf{+38.6} & +22.2 & +28.0 & \textbf{+43.6} \\
  & NPO     & +15.7 & \textbf{+15.9} & \textbf{+20.3} & +11.3 & \textbf{+16.5} & +11.1 & +13.0 & \textbf{+18.1} \\
  & GAGDR   & +23.8 & \textbf{+30.7} & \textbf{+28.9} & +25.7 & \textbf{+26.6} & -11.1 & +15.0 & \textbf{+24.5} \\
  & GA      & +16.9 & \textbf{+19.9} & \textbf{+21.3} & +15.5 & \textbf{+24.1} & +13.9 & +21.0 & \textbf{+23.4} \\
\bottomrule
\end{tabular}%
}
\caption{Hold-out-language drop rates per unlearning method for $p5$. \textbf{Bold} marks the higher value within each comparison pair (High/Low and Lat/nLat). G and Q denote Gemma3 and Qwen2.5, respectively.}
\label{tab:unlearn_ho_groups_p5}
\end{table*}



%

\end{document}